\documentclass{article}

    \usepackage[final]{neurips_data_2023}

\PassOptionsToPackage{hyphens}{url}
\usepackage[utf8]{inputenc} %
\usepackage[T1]{fontenc}    %
\usepackage{hyperref}       %
\hypersetup{breaklinks=true,colorlinks,citecolor=gray,linkcolor=blue,urlcolor=blue}
\usepackage{url}            %
\usepackage{booktabs}       %
\usepackage{amsfonts}       %
\usepackage{nicefrac}       %
\usepackage{microtype}      %
\usepackage[dvipsnames]{xcolor}         %

\usepackage{amsmath}
\usepackage{cleveref}
\usepackage{graphicx}
\usepackage{listings}
\usepackage{colortbl}
\usepackage{soul}

\usepackage[font=small,labelfont=bf]{caption}
\title{AI for Scaling Legal Reform: Mapping and Redacting Racial Covenants in Santa Clara County}

\author{%
\begin{tabular}{ccc}
\begin{tabular}[t]{c}
Faiz Surani\thanks{Equal contribution.} \\
{\normalfont Stanford University} \\
\end{tabular} &
\begin{tabular}[t]{c}
Mirac Suzgun\footnotemark[1] \\
{\normalfont Stanford University} \\
\end{tabular} &
\begin{tabular}[t]{c}
Vyoma Raman \\
{\normalfont Stanford University} \\
\end{tabular} \\ \addlinespace[4ex]
\begin{tabular}[t]{c}
Christopher D. Manning \\
{\normalfont Stanford University} \\
\end{tabular} &
\begin{tabular}[t]{c}
Peter Henderson \\
{\normalfont Princeton University} \\
\end{tabular} &
\begin{tabular}[t]{c}
Daniel E. Ho\thanks{Corresponding author: \url{deho@stanford.edu}.} \\
{\normalfont Stanford University} \\
\end{tabular}
\end{tabular}
\\ \\
 { \emph{Warning: This paper contains offensive racial terms from actual deed documents.}}
}

\begin{document}

\maketitle
\setcounter{footnote}{0} 

\begin{abstract}
Legal reform can be challenging in light of the volume, complexity, and interdependence of laws, codes, and records. One salient example of this challenge is the effort to restrict and remove racially restrictive covenants, clauses in property deeds that historically barred individuals of specific races from purchasing homes. Despite the Supreme Court holding such racial covenants unenforceable in 1948, they persist in property records across the United States. 
Many jurisdictions have moved to identify and strike these provisions, including California, which mandated in 2021 that all counties implement such a process. Yet the scale can be overwhelming, with Santa Clara County (SCC) alone having over 24 million property deed documents, making purely manual review infeasible. We present a novel approach to addressing this pressing issue, developed through a partnership with the SCC Clerk-Recorder's Office. First, we leverage an open large language model, fine-tuned to detect racial covenants with high precision and recall. We estimate that this system reduces manual efforts by 86,500 person hours and costs less than 2\% of the cost for a comparable off-the-shelf closed model.  Second, we illustrate the County's integration of this model into responsible operational practice, including legal review and the creation of a historical registry, and release our model to assist the hundreds of jurisdictions engaged in similar efforts. Finally, our results reveal distinct periods of utilization of racial covenants, sharp geographic clustering, and the disproportionate role of a small number of developers in maintaining housing discrimination. We estimate that by 1950, \emph{one in four} properties across the County were subject to racial covenants. 
\end{abstract}

\clearpage

\section{Introduction}

Legal reform is complex. When a court declares a statutory provision unconstitutional, a legislature prohibits a certain practice, an agency initiates regulatory review, or the public adopts a referendum, such changes can ripple through thousands of code provisions, a thicket of regulations, and millions of administrative records. 
Armies of lawyers and clerks can spend thousands of hours to identify legal dependencies to implement such changes. Because this process is so resource-intensive, many outdated legal provisions can persist in official documents for decades.

One prominent example of this issue is the persistence of racially restrictive covenants in real property deeds. Examined extensively by lawyers and social scientists \citep{brooks2011covenants, brooksrose2013saving,gonda2015unjust,gotham2000urban,jones2000origins,ming1949racial,roisman2022stumbling,rothstein2017color,rose2016racial, rose2022property,rose2023general, vose1967caucasians,yalereport}, racial covenants are discriminatory clauses that prohibited the purchase, lease, or occupation of land based on race.\footnote{Discriminatory restrictive covenants may also apply to other attributes, such as religion, family status, and national origin, but we focus on the principal case of racial covenants here.} Although declared \emph{unenforceable} by the United States Supreme Court in \emph{Shelley v.\ Kraemer} (1948)\footnote{334 U.S.\ 1.} and \emph{illegal} under the Fair Housing Act (1968),\footnote{42 U.S.C.\ \S~3600, et seq.} such covenants continue to exist in the pages of real property records across the United States. The sheer volume makes identifying and redacting racial covenants both monumental and resource-intensive \citep{howard2021california}. Over a dozen  jurisdictions have enacted legislation to address racial covenants \citep{yalereport}. The typical approach has been to 
enable individual homeowners to petition for legal review to redact these records to limited effect.\footnote{\citet[][noting from interviews with county clerks that ``only a few people have used these provisions'']{yalereport}.}

But change is afoot. In 2021, California enacted Assembly Bill 1466 \citep[AB 1466;][]{california_ab1466},\footnote{AB 1466 is codified at Cal.\ Gov.\ Code \S~12956.3.} which mandates that all 58  counties develop programs to affirmatively identify and redact racial covenants from property records.\footnote{Under California's Fair Employment and Housing Act, unlawful restrictions are also ones based on age, race, color, religion, sex, gender, gender identity, gender expression, sexual orientation, familial status, marital status, disability, veteran or military status, national origin, ancestry, genetic information, or source of income. Cal.\ Gov. Code \S~12955.}  While AB 1466 is seen as an important step to recognizing and mitigating the remnants of institutionalized housing discrimination, its implementation presents significant challenges. Santa Clara County, for example, has more than 24 million property records, spanning over 84 million pages, including some that date back to the 1850s.\footnote{Santa Clara County Clerk-Recorder's Office, Restrictive Covenant Modification Program Implementation Plan, 
\url{https://clerkrecorder.sccgov.org/unlawfully-discriminatory-restrictive-covenant-modification-program-assembly-bill-1466}}

The complexity and scale of these historical documents, some of which are handwritten or stored on decades-old microfiche cards, render manual review infeasible. Given available resources for manual review, it could take a single county about 160 years and over \$22 million to complete a scan of all 24 million records.\footnote{At California's 2024 minimum wage of \$16.00 per hour, the projected cost of hiring human reviewers to manually examine the Santa Clara County's entire real property records would be about \$22.4 million.}

In a unique multiyear partnership between Stanford RegLab and the Santa Clara County Clerk-Recorder's Office, along with a collaborator at Princeton University, we prototyped, developed, and operationally integrated a machine learning-based pipeline to identify and map racial covenants at scale. 
Our system is capable of processing millions of documents in a single day. It offers an efficient and reliable solution that drastically reduces the time and labor required for manual review. As shown in Table~\ref{tab:cost_comparison}, our machine-learning pipeline has saved over 86,500 hours of manual human labor, costing less than 0.02\% of a full manual human review and under 2\% of a comparable off-the-shelf-model such as OpenAI's ChatGPT-3.5. Our solution offers an accurate, fast, and cost-effective  path for Santa Clara County -- and other jurisdictions -- to best utilize limited human resources, meet legislative requirements, and preserve important historical records for further study.

This article presents three contributions stemming from this effort. \emph{First}, we present our machine-learning pipeline for identifying and mapping racial covenants. Our pipeline processes images of historical property deeds, converts them into text, and then leverages a state-of-the-art, finetuned language model to accurately detect racial covenants. If unlawful language is found in a deed, the system highlights the content and extracts the property address. Both the highlighted documents and their corresponding address are then sent to Santa Clara County for legal review and final confirmation. Remarkably, our model achieves extremely strong performance, with a precision of 1.0 and a recall of 0.99 on an evaluation suite of real property deeds. 

We show that this AI-based approach offers significant advantages over traditional methods, such as keyword-based searches, which are prone to substantial false-positive rates. Scanning artifacts, such as poor OCR quality in older deeds, and ambiguous terms like ``white'' (which could refer to a person's name or a street) contribute to the inaccuracies of lexical search techniques. In contrast, our system analyzes the full semantic context of each document, enabling it to detect racial covenants that have atypical language structures or obscure phrasing, some of which had previously gone unnoticed by manual reviewers.

\emph{Second}, we discuss how we integrated the model into a responsible operational process that includes thorough legal review and the creation of a historical registry of the removed racial covenants. By retaining a historical record, we ensure that this dark chapter of housing discrimination is not erased from public memory, but preserved and understood. We also make our models, results, and web interface for reviewing records available to assist the hundreds of jurisdictions engaged in similar efforts to identify these unenforceable legal provisions.\footnote{These will be made available at \url{https://reglab.github.io/racialcovenants/}.} 

\begin{figure}[t]
    \centering
    \includegraphics[width=\linewidth]{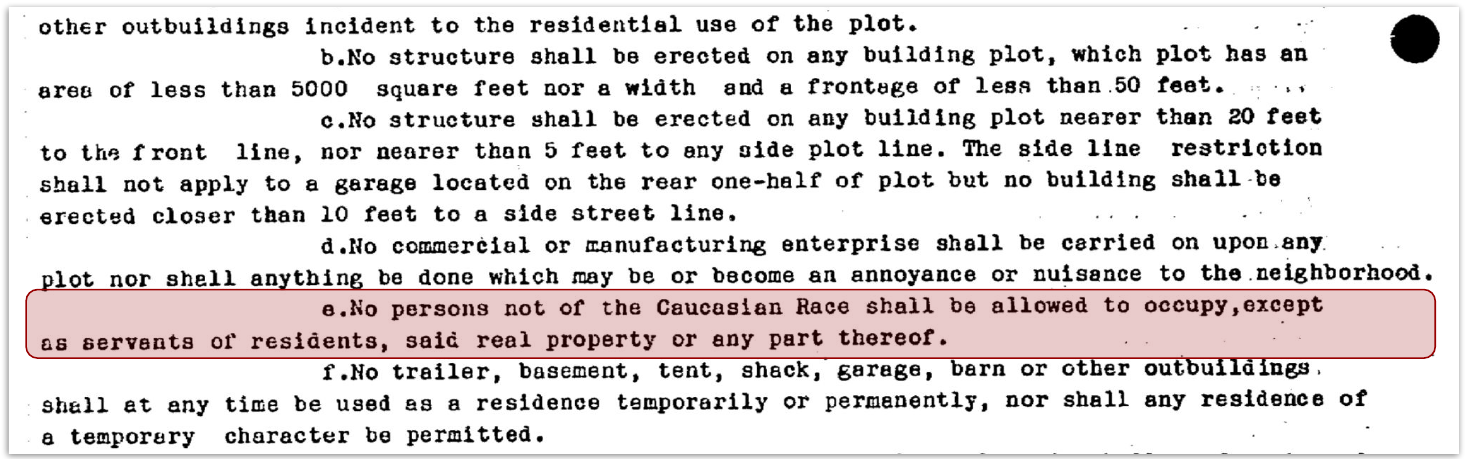}
    \caption{Although racially restrictive covenants are no longer legally enforceable and are considered illegal under the Fair Housing Act today, they still exist in thousands, possibly even millions, of historical property records in California. One such example, found in a 1940 real property deed from Santa Clara County's archives, contains the following discriminatory language: ``{No persons not of the Caucasian Race shall be allowed to occupy, except as servants of residents, said real property or any part thereof.}'' The deed further specifies that ``[t]hese covenants are to run with the land and shall be binding on all parties,'' thereby affecting not only the tenants at the time but also the potential future owners of the land.}
    \label{fig:example}
\end{figure}

\emph{Third}, our findings shed light on the history of racial exclusion in the California housing market. The racial covenants identified by our system reveal distinct patterns of racial categorization, usage across time, and geographical clustering, adding to an important body of scholarship on housing discrimination and racial covenants. Our large-scale dataset enables researchers to understand and test for different accounts of racial covenants. Consistent with existing accounts \citep{rose2016racial}, early racial covenants in California specifically focused on Asian groups, but the number of covenants expressly barring black homeowners was at the same rate in the early 20th century, even when Asian residents far outnumbered black residents. We observe a drop in racial covenants after \emph{Shelley}, but consistent with \citet{brooksrose2013saving}, racial covenants persist well after 1948. We also find that a state actor (the city of San Jose) owned land subject to a substantial number of burial deeds with racial covenants (e.g., burial plots exclusively for ``Caucasian race''), complicating conventional accounts of covenants as a private substitute for public state action (racial zoning) found unconstitutional in 1917. Just ten developers appear responsible for nearly a third of racial covenants in the County, suggesting more \emph{agency} in the construction of what became Silicon Valley \citep{howell,redford, taeuber1961privately}.  We provide an estimate that \emph{one in four} properties the county were covered by a racial covenant in 1950. 

Overall, our project demonstrates the power of machine learning and large language models to play a substantial role in scalable legal reform and the public sector \citep{engstrom_government_2020}. At core, we show how AI can meaningfully assist to unearth, preserve, and shine a light on housing discrimination in a way that was obscured by previously inaccessible deed records.\footnote{As we articulate below, California deed records are not publicly available at scale, despite being public records.} 

\textbf{Organization}. Our paper proceeds as follows. Section~\ref{sec:background} provides background on racial covenants, California's reform efforts, and existing efforts to map and redact deeds. %
Section~\ref{sec:data} discusses the data processing steps to digitize, augment, and label deed records for machine learning. Section~\ref{sec:pipeline} discusses the AI-based detection pipeline, which shows that large language models enable substantial improvements over keyword-based searches, and the geolocation of records. Section~\ref{sec:evaluation} presents results, which show remarkable improvements, such as the reduction in the false positive rate from 28.9\% with keyword searches to 0\% with a fine-tuned open-source language model. Section~\ref{ref:integration} discusses how we integrated the AI system to preserve legal review of each redacted provision, but with dramatically lower search costs. Section~\ref{sec:evolution} shows how this comprehensive effort enables us to unearth rich historical facts about the evolution of racial covenants in Santa Clara County. Section~\ref{sec:prevalence} presents an estimate of the proportion of 1950 housing stock that was covered by racial covenants. Section~\ref{sec:limits} discusses limitations of the approach and Section~\ref{sec:conclude} concludes. 

\begin{table}[!t]
\centering
\begin{tabular}{l l l}
\toprule
\textbf{Method} & \textbf{Time} & \textbf{Monetary Cost} \\
\midrule
\textbf{Manual Review} {\footnotesize(One Staff Member)} & 9.89 years & \$1,400,000   \\
\textbf{Off-the-Shelf LM} {\footnotesize(GPT-3.5)} & 3.63 days & \$13,634   \\
\textbf{Off-the-Shelf LM} {\footnotesize(GPT-4 Turbo)} & 3.63 days & \$47,944   \\
\textbf{Our Custom LM} {\footnotesize(Finetuned Mistral)}  & 6 days & \$258 \\
\bottomrule
\end{tabular}
\vspace{0.2em}
\caption{Resource cost comparison for identifying racially restrictive covenants in Santa Clara County’s 5.2 million pages of property records from 1907 to 1978. Our custom finetuned Mistral model stands out as the most scalable and economical solution, completing the full review in just six days for \$258, a small fraction of the cost and time required for manual review, which would cost over \$1.4 million and take years to finish. For additional details, please refer to Section~\ref{sec:resource-cost-comparison} in the Appendix. %
}
\label{tab:cost_comparison}
\end{table}

\section{Background}
\label{sec:background}

\subsection{Racial Covenants}
Racially restrictive covenants were legal clauses embedded in property deeds that prohibited the sale, lease, or occupation of land by individuals based on race. These covenants became a widespread tool for enforcing residential segregation in the United States during the first half of the 20th century. Generally, covenants ``\emph{run with the land},'' which meant that restrictions affect not only current but also all future owners of the real property~\citep{brooksrose2013saving}. While African Americans were the primary targets of racial covenants, other groups, such as Asians, Latinos, Jews, and Southern and Eastern Europeans, were also excluded from certain neighborhoods through the use of these discriminatory binding clauses~\citep{brooksrose2013saving}. Racial covenants were designed to maintain racially homogenous, white-majority neighborhoods by barring minority groups from settling in specific areas, and they were actively supported by real estate boards, developers, homeowner associations, and governmental institutions \citep{brooksrose2013saving}.

Racial covenants originated in the mid-19th century, but became more prevalent after  the Supreme Court held racial zoning unconstitutional in \emph{Buchanan v.\ Warley}.\footnote{245 U.S.\ 60 (1917). See also \citet[78]{rothstein2017color}.} During this period, white homeowners viewed racial covenants as a means to protect property values and maintain racial homogeneity within their communities. Integrating neighborhoods with non-white residents, particularly African Americans, was perceived as leading to economic decline and social instability.\footnote{\href{https://fraser.stlouisfed.org/files/docs/publications/fha/1936apr_fha_underwritingmanual.pdf}{The April 1936 edition of the \emph{Underwriting Manual} of the Federal Housing Administration} explicitly stated this as a policy as follows: 
``If a neighborhood is to retain stability it is necessary that properties shall continue to be occupied by the same social and racial classes. A change in social or racial occupancy generally leads to instability and a reduction in values. The protection offered against adverse changes should be found adequate before a high rating is given to this feature. Once the character of a neighborhood has been established it is usually impossible to induce a higher social class than those already in the neighborhood to purchase and occupy properties in its various locations.'' (Part II, para. 233)} Consequently, racial covenants were often marketed as desirable features in new suburban developments, which promoted ``restricted'' neighborhoods as more valuable, secure, and exclusive~\citep{santucci2020documenting,rothstein2017color}.

A typical restrictive covenant had the following language: 

{\setlength{\leftmargini}{2em} %
 \begin{quote}
 \footnotesize{
       ``No part of said property shall be sold, let, or leased, transferred, or assigned to, or occupied by any person not of the Caucasian race, or to be used by any other than a person of the Caucasian race.''
       }
 \end{quote}
}
But significant variation in the precise language is known to exist. 

Real estate institutions like the National Association of Real Estate Boards (NAREB) played a crucial role in institutionalizing racial covenants. From 1924 to 1950, NAREB’s code of ethics required realtors to engage in racial steering provisions, effectively ensuring that minority buyers were not introduced into white neighborhoods~\citep{brooksrose2013saving}.\footnote{The Article 34 of Part III of the 1924 NAREB's Realtors' Code of Ethics stated: ``A Realtor should never be instrumental in introducing into a neighborhood a character of property or occupancy, members of any race or nationality, or any individuals whose presence will clearly be detrimental to property values in that neighborhood.'' (as quoted in~\citep{brooksrose2013saving}.)} 
Violating this code could result in expulsion, further promoting racial segregation within the real estate industry \citep{santucci2020documenting}. Federal programs such as the Federal Housing Administration further entrenched these practices by making racial covenants a condition for mortgage insurance approval, thus embedding racial segregation in housing markets across the country~\citep{brooksrose2013saving}.

In 1948, the Supreme Court found racial covenants to be unenforceable in \emph{Shelley v.\ Kraemer}. The federal Fair Housing Act of 1968 prohibited the use of racial covenants. But because  covenants run with the land, such provisions have remained on the books, even if unenforceable. Much debate exists around the persistent impact of racial covenants. One perspective is that covenants institutionalized segregation in the housing market, contributing to enduring racial disparities in wealth accumulation, homeownership, and access to essential resources such as education and public services~\citep{MappingPrejudice2022_RRCs}. \citet{brooksrose2013saving} argue that racial covenants continued to have effect post-\emph{Shelley} as \emph{signaling} devices for the kind of community associated with the property. They provide an account stemming largely from litigated cases and the history in Chicago, and argue, based on game theory, that racial covenants were most widely deployed in ``loosely knit'' communities requiring a signaling device (id.). Using neighborhood data from Chicago, \citet{brooks2011covenants} estimates that racial covenants had effects lasting past \emph{Shelley}, consistent with signaling. Yet because a comprehensive register of racial covenants is so difficult to compile, studies have been limited in their ability to examine or test these accounts with quantitative evidence about the prevalence, dynamics, and geographic correlates of racial covenants.\footnote{\citet{brooks2011covenants}, for instance, provides one of the few quantitative analyses, but had to rely on covenant data at the neighborhood level.}

\begin{figure}[t]
    \centering
    \includegraphics[width=\linewidth]{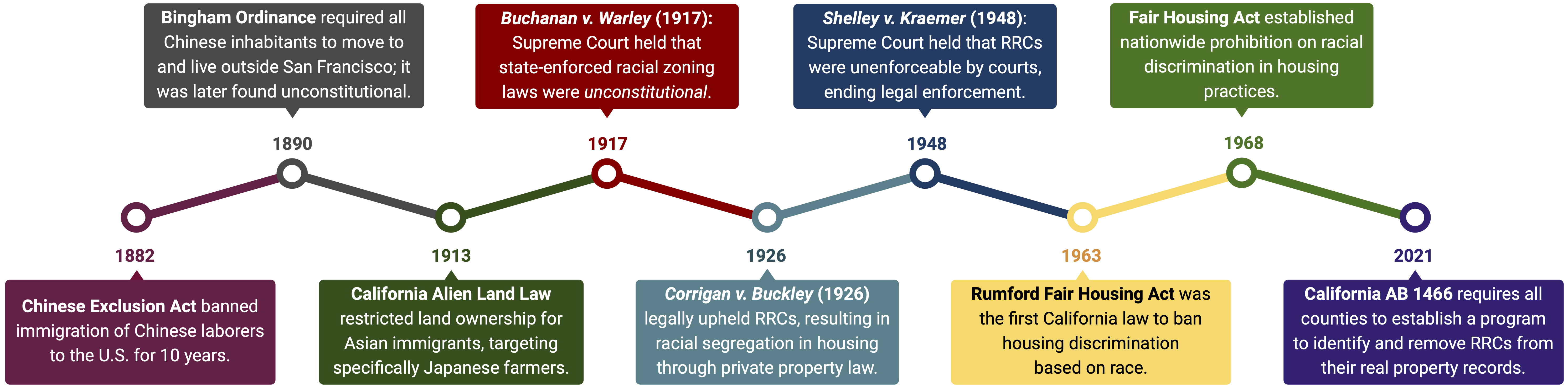}
    \caption{Brief overview of legal developments that impacted California's housing market in the 20th century. The Rumford Act was overturned by Proposition 14, which was in turn found unconstitutional by the California Supreme Court in \emph{Mulkey v.\ Reitman}, 64 Cal. 2d 529 (1966).}
    \label{fig:overall}
\end{figure}

\subsection{California Legislation}

In recent years, there has been growing desire to address the ongoing presence of racial covenants in property records. Numerous states, including California, Washington, Minnesota, and Texas, have passed laws allowing property owners to remove racial covenants from their deeds.  However, these laws typically place the responsibility on individual homeowners, resulting in a piecemeal approach. Between 1999-2021, California, for instance,  maintained a process by which homeowners could petition with the County Recorder to modify a racial covenant on their property. \citet{brooksrose2013saving} note that a homeowner would have to be quite ``dedicated'' to pursue that process, as the disclosure of the process is given to purchasers along with all other home disclosures.  As in other jurisdictions that put the onus on individual homeowners \citep{yalereport}, Santa Clara County had only received a handful of such requests prior to 2021.\footnote{Cal. Gov. Code \S\S~12956.1-12956.2.} One state legislator opined, ``the present system
is underutilized and public awareness on the issue is low.'' \citep{howard2021california} 

In recognition of this limitation, California adopted a more proactive and  comprehensive approach with the passage of Assembly Bill 1466 \citep{california_ab1466}. Instead of relying on homeowner initiative, AB 1466 mandates that all 58 counties in California establish programs to identify and redact racial covenants from property records. In addition, AB 1466 mandates the retention of each ``nonredacted record for future reference and public request needs.''\footnote{Cal.\ Gov.\ Code \S 12956.3(c).} In short, counties must redact unlawful discriminatory language in active property records, while retaining historical deed records.\footnote{Though, there is some leniency for false positives or false negatives: ``The failure of a county recorder to identify or redact illegal restrictive covenants, as required by this section, or the county recorder’s identification or redaction of any restrictive covenants that are later determined not to be illegal, shall not result in any liability against the county recorder or the county.'' Cal.\ Gov.\ Code \S 12956.3(g).}

The implementation has posed serious logistical challenges. A prior proposal faced resistance by county recorders for ``creat[ing] an enormous
workload'' and posing a ``potential near shut-down of county recorder offices.''\footnote{Prior proposals (AB 2204 and AB 985) would have placed the responsibility on title companies \citep{howard2021california}.} 
First, because these records span all county properties and all historical transactions, the sheer record volume is large. In counties such as Santa Clara -- with some 24 million of property records -- purely manual review is impractical. Second, the language used to identify racial groups and prohibitions can vary substantially over time and place.  Third, AB 1466 requires review by county counsel to formally record any amendment, making the process organizationally challenging to navigate. Fourth, while AB 1466 included a fee provision that allocated \$2 of recording fees per specified document for funding the implementation of AB 1466, such fees may not cover the full costs.\footnote{This fee provision would lapse by December 31, 2027.} In Santa Clara County, this would fund up to three positions, excluding costs for time for review by the county counsel, digitization of records, and any software to aid in the process. 
Some counties have turned to third-party private vendors to expedite the process. Los Angeles, for instance, hired a private firm in an \$8 million contract to carry out the process over seven years.\footnote{Jaclyn Cosgrove, Racist History Lives on in Millions of Housing records. L.A. County is about to fix that, \emph{L.A.\ Times}, Feb.\ 6, 2024,  \url{https://www.latimes.com/california/story/2024-02-06/l-a-county-will-remove-racist-restrictive-covenant-language-from-millions-of-documents}.} But resources across counties can vary widely. In late 2022, Santa Clara piloted a manual review process with  two staff members manually reviewing 89,000 pages of deeds, finding roughly 400 racial covenants. At an average of a minute per page, performing a manual review of the entire collection of 84 million pages of records would require approximately 1.4 million staff-hours, amounting to nearly 160 years of continuous work for a single individual.\footnote{Please refer to Section~\ref{sec:resource-cost-comparison} in the Appendix for additional details about our assumptions and calculations.} As \citet{howard2021california} notes, the big open question is whether ``fifty-eight different county recorders
in California are able to develop and maintain redaction
procedures that are consistent, predictable, effective,
efficient, and easily implemented.''

\subsection{Existing Efforts}

Outside of California, there are also increasing efforts to identify and address racial covenants. 

First, efforts at the national level have grappled  with the resource-intensive nature to identify racial covenants. The Uniform Law Commission proposed model legislation vesting principal responsibility in individual homeowners, mimicking California's pre-2021 process.\footnote{\url{https://www.uniformlaws.org/committees/community-home?CommunityKey=b1ed931f-d4c2-4078-867d-018a850ef303}} On the other hand, the federally proposed Mapping Housing Discrimination Act would provide grants to educational institutions to ``analyze, digitize, and map historic housing discrimination'' with a goal of a national database of racial covenants.\footnote{\url{https://www.congress.gov/bill/117th-congress/senate-bill/2549}.} The question of who bears responsibility hinges critically on understanding scalable approaches to sifting through deed records, as well as ongoing California efforts. 

Second, academic and grassroot initiatives have crowd-sourced efforts through large numbers of volunteers \citep{bakelmun2019open}. 
The University of Minnesota's \href{https://mappingprejudice.umn.edu}{\textit{Mapping Prejudice}} project, for instance, was one of the earliest  initiatives and relied on thousands of community volunteers to manually sift through digitized property deeds in Hennepin County, Minnesota, home to the city of Minneapolis. The \href{https://www.chicagocovenants.com/}{\textit{Chicago Covenants Project}} took a metropolitan-wide approach to documenting a range of historical housing practices, including racial covenants.  The \href{https://www.justiceindeedmi.org/}{\textit{Justice InDeed}} project similarly focused on identifying racial covenants and collaborating with community stakeholders to raise awareness and pursue local solutions in Washtenaw County, Michigan. 
These participatory approaches are laudable for improving community understanding of the local history of housing discrimination. The reliance on volunteers to sift through an immense volume of records, however, can make this approach infeasible for all jurisdictions.\footnote{A network of these initiatives was formed as the ``\href{https://www.nationalcovenantsresearchcoalition.com/}{National Covenants Research Coalition}.'' 
} 

Third, other projects have been state-initiated. Over a dozen states have enacted legislation to address racial covenants \citep{yalereport}. In 2022, Washington state, for instance, mandated a review of property records across the state to identify racial covenants.\footnote{Concerning Review and Property Owner Notification of Recorded Documents with Unlawful Racial Restrictions, SHB 1335 (2021), codified at Wash.\ Rev.\ Code Ann.\ \S~49.60.525 (2021).} This led to the establishment of the \href{https://inside.ewu.edu/racial-covenants-project/}{\textit{Eastern Washington Racial Covenants Research Project}}, supported by universities and state agencies. While more institutionalized, the manual nature of the review process still remains resource-intensive.\footnote{See Samantha Wohlfeil, ``As EWU readies to share maps of racial covenants in Eastern Washington, a Spokane title company is helping homeowners disavow the racist documents,'' \emph{Inlander}, April 25, 2024 (``[P]rofessors and student employees traveled to auditors' offices to dig through deed books by hand.'').}

Each of these efforts represents an important initiative in understanding the practice and history of housing discrimination. These initiatives have mobilized public interest, enlisted volunteers, and raised grassroots awareness of racial covenants. Yet the core shared approach of relying on purely human review can slow the rate at which jurisdictions can learn about these records and their impact on local histories and be infeasible for many other jurisdictions. Numerous California counties, for instance, have aimed to complete a scan of racial covenants by 2027 and simply do not have scores of volunteers or staff to peruse records. The legal obligation is placed in the offices of the county recorder and counsel, requiring a comprehensive, systematic, and scalable solution to prioritize limited human resources. As far as we are aware, no prior efforts have explored the power of large language models to assist in this process.\footnote{\textit{Mapping Prejudice} has used forms of automation to transcribe and analyze documents, but not modern machine learning in the search process itself.}

\section{Data} 
\label{sec:data}

We now describe the data used for training and deploying our AI-based approach. 
Gathering this data required a significant digitization process, extracting and processing 5.2 million pages of deeds stretching back to the 1850s (\S~\ref{sec:digit}). We then supplemented this data with historical deed records available online from around the country (\S~\ref{sec:otherdata}), which has the coincidental benefit of enabling us to assess the model's robustness across jurisdictions.  Finally, we manually annotate 3,801 deeds to build a training dataset and held-out evaluation dataset for our AI pipeline (\S~\ref{sec:annotation}).

\subsection{Digitization, Collection, and Sharing of Real Property Deeds}
\label{sec:digit}
The Santa Clara County Clerk-Recorder's Office has an extensive archive of over 24 million real property deeds. Of these records, approximately 18 million -- issued since 1980 -- are stored digitally, while the remaining 6 million deeds -- created before 1980 -- were originally preserved on physical microfiche sheets. More than a decade prior to our work, the County had engaged a vendor to scan these records into a proprietary system known as Digital Reel; however, as we discuss in Appendix \ref{appendix_ocr}, the quality of these scans was poor and required significant post-processing.

Our partnership around exploring the use of AI began in October 2022. One of the notable barriers to transparency around deed records in California lies in a statutory mandate to charge fees for any copies of recorded documents.\footnote{Cal.\ Gov.\ Code \S~27366 provides: ``The fee for any copy of any other record or paper on file in the office of the recorder, when the copy is made by the recorder, shall be set by the board of supervisors in an amount necessary to recover the direct and indirect costs of providing the product or service or the cost of enforcing any regulation for which the fee or charge is levied.'' This provision has been subject to extensive litigation. See, e.g., California Public Records Research, Inc.\ v.\ County of Stanislaus, 246 Cal.\ App.\ 4th 1432 (2016);  California Public Records Research v.\ County of Yolo, 4 Cal.\ App.\ 5th 150 (2016).}  In other words, despite their status as public records, deed documents are available only on an individual fee basis. Given the massive scale of the review task, purchasing deed records would, of course, have been prohibitively expensive.\footnote{At a cost of \$4 for the first page and \$2 for each subsequent page, purchasing the 5.2M pages (with the average deed running 2.5 pages) might have cost over \$13 million.} Through our partnership, we developed unique a data sharing agreement, enabling the Stanford team  to process deed data, with the County retaining ownership of the records.

We began our work on samples of 20,000 pages of property deeds filed between 1900 and 1940, manually exported from the County's Digital Reel system. This 20,000-page sample enabled us to rapidly develop and refine our automated detection pipeline. 

After this piloting phase, the County extracted the full collection of pre-1980 scans in February 2024. This represents roughly 5.2 million pages of real property documents from 1865 to 1980. We focus our analysis on documents from 1902 to 1980 for two reasons. First, deeds filed prior to 1902 were handwritten rather than typed, and we found no available OCR tools to be effective at transcribing these documents.\footnote{We did explore developing a bespoke computer vision or multimodal text-vision system.} Second, records after 1980 contain protected fields like Social Security information, so we avoided ingesting sensitive data and potentially training our model on it, which may have raised privacy and legal concerns. As we note above and consistent with our results in Section~\ref{sec:evolution}, 1902 to 1980 likely covers the vast majority of racial covenants in the County; the first racial covenant we find was filed in 1907 and the last in 1974.

\subsection{Data Augmentation}
\label{sec:otherdata}
Both within California and across the nation, historical property deeds vary significantly in format, phrasing, and, when digitized, OCR quality. In order to build a system that is robust to these variations, we supplemented the Santa Clara County dataset with property deeds from around the nation, both with and without racial covenants.

Since property records in California counties are not freely accessible, we expanded our search to other counties in the United States. Using \href{https://govos.com/products/public-record-access/channel/}{GovOS Cloud Search}, we identified seven counties whose ``Official Records Search’’ platforms allowed users to freely search and download real property deeds, although downloads were limited to fifty records per batch.\footnote{\url{https://kofilehelp.zendesk.com/hc/en-us/sections/4416665864343-Cloud-Search-Active-Sites}.} These platforms enabled searches by metadata and keyword terms. To gather a seed dataset of deeds with a high probability of containing racial covenants, we conducted manual searches for terms typically associated with such covenants, such as ``No person of,'' ``Caucasian,'' ``Negro,'' and other relevant racial terms. This method provided us with more than 10,000 property deeds from seven counties: Bexar County, Texas; Cuyahoga County, Ohio; Denton County, Texas; Franklin County, Ohio; Hidalgo County, Texas; and Lawrence County, Pennsylvania. This approach not only helped us collect relevant data but also allowed us to assess the generalizability of our model across different jurisdictions. As we discuss below, we specifically investigate the limitations of keyword-based approaches, and find that context-aware language models boost performance substantially.

\subsection{Annotation} \label{sec:annotation}

We labeled our data collection by identifying quotes that contain racial covenants on each page. This annotation occurred over three stages: initial training data generation, model prediction review, and rich annotation.

In our initial round of annotation, we selected a sample of 3,000 pages in our collection based on keywords that almost certainly indicate the presence of a racial covenant in the deed text. These include terms like ``Negro,'' ``Mongolian,'' and ``Asiatic.'' We partnered with data annotation company CloudFactory to help us identify and label racial covenants in these pages.\footnote{During our collaboration with CloudFactory for data annotation, we carefully prepared comprehensive documentation to guide the annotators through the task. Given the potentially sensitive nature of the material—historical property deeds containing racially restrictive covenants, as well as accounts that could be considered offensive or harmful to some readers—we issued a clear advisory to approach the content with care. We emphasized that the annotators could stop the task at any point if they felt uncomfortable. In addition, we consulted with CloudFactory’s management to ensure that appropriate counseling and support resources would be available to their team, should any annotator feel the need for assistance or support. Our priority was to handle this material with the utmost sensitivity, while ensuring the well-being of those involved in the annotation process.}

After training models and generating predictions, we reviewed their performance. For all positive predictions, we labeled whether they were true positives or false positives. These ensured that we verified the small number of positive examples as well as hard negative examples. We additionally sampled and verified negative predictions to ensure some balance in the data. These new annotations were incorporated into the training set of future models.

Recognizing the need to easily validate model predictions and locate racial covenants on a page, we built a web application to assist with rich annotation. This made it easy to precisely select a bounding box on the image of the deed book page and compute a text span for the annotation process, while simultaneously allowing us to visualize predictions for verification.

All combined, including both Santa Clara County documents and documents from across the country, we collected 3,801 annotations of deed pages, of which 2,987 (78.6\%) contained a racially restrictive covenant. Notably, this annotation requires human review, but at a much smaller scale than reviewing all records.

\section{An AI-based Racial Covenant Detection System}
\label{sec:pipeline}

Our racial covenant detection pipeline first takes deed images, converting them to text through optical character recognition (OCR) machine learning models. We pre-process and clean all of these documents (\S~\ref{sec:preprocessing}).
Next, we process this text to detect racial covenants. We test five different approaches for this, including a custom finetuned large language model (\S~\ref{sec:detection}). Then we identify the part of the original image containing the racial covenant, highlighting it for Santa Clara County officials (\S~\ref{sec:span}). Finally, we extract any geographic information from the deed to help map the location of the racial covenant (\S~\ref{sec:geo}). 

\subsection{Preprocessing and OCR}
\label{sec:preprocessing}
Our data preprocessing pipeline operates on property deeds split at the page level. We run OCR to retrieve the text of each page. We used the docTR OCR library~\citep{doctr2021} to run a VGG16 model~\citep{simonyan2014very} for text recognition and a DB-ResNet-50 model~\citep{liao2020real} for text detection.

\subsection{Racial Covenant Detection}
\label{sec:detection}

Once converted to text, we test five approaches for identifying and extracting racial covenants: keyword matching, fuzzy keyword matching, zero-shot GPT-3.5 Turbo, few-shot GPT-3.5 Turbo, and a custom finetuned Mistral 7B parameter open source model.

\textbf{Keyword matching.} The rudimentary approach approximates keyword searches that humans may conduct on digitized text, and follows the baseline that the county had established.  The approach matches historically common racial or ethnic terms in the text, such as ``Caucasian,'' ``Mongolian,'' or ``Negro.'' We implemented a simple substring matching approach to detect these keywords. We construct this list by first conducting a manual review of known racial covenants and then manually adding additional variations. We also compared this list to a list of terms compiled by the County, which was comparable.\footnote{The full list of keywords used by the County is included in Section~\ref{sec:scc-keyword-list} of the Appendix.}

\textbf{Fuzzy Keyword Matching.} Keyword matching identifies some racial covenants, but OCR errors can result in many false negatives. For instance, the term ``Caucasian'' was sometimes transcribed as ``Caucian'' or ``Causasan'' on scans with significant visual artifacts. To address these issues, we implement a simple augmentation to keyword matching: {fuzzy keyword matching}.

Our fuzzy matching approach first tokenizes deed text into words. Then, each word is broken down into a set of trigrams. We then compute the cosine similarity between the set of trigrams in each word of the deed and the trigrams of each keyword in our list. We conclude a deed contains a keyword if the maximum cosine similarity between any word in the deed and any keyword in our list is greater than the threshold value of 0.75.

\begin{figure}[t]
    \centering
    \includegraphics[width=\linewidth]{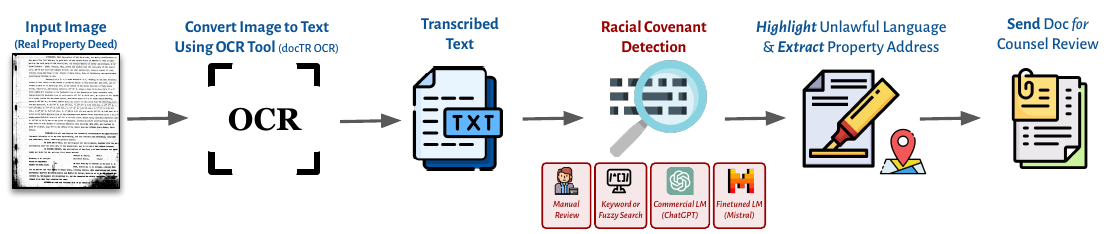}
    \caption[]{\emph{Diagram of our pipeline for detecting racial  covenants.} The process begins by converting an image of a property deed into text using an OCR tool (docTR). The transcribed text is then analyzed for racially discriminatory language. If such unlawful language is found, the system highlights the content and extracts the property address. Both the highlighted language and the corresponding address are sent then to Santa Clara County for legal review and final confirmation.}
    \label{fig:annual_rrc_numbers}
\end{figure}

\textbf{Large Language Models.} Lexical approaches -- while computationally inexpensive and fast -- are still inherently limited by their lack of contextual understanding of the textual content. Both lexical approaches described above require the pre-specification of a set of keywords to match. Even minor perturbations of the term list can lead to significant changes in output quality. For instance, the term ``white'' is a key phrase used in many racial covenants, but those usages are far outnumbered by innocuous uses of the term, such as in street names, surnames, descriptions of an object's color, and other contexts. In addition, the specific terms commonly used in racial covenants vary significantly by time period and location.

To address this, we design detectors based on large language models (LLMs), which have shown promise in other natural language generation and understanding settings. We test three versions.

First, we evaluate the closed-weight GPT-3.5 Turbo model with a zero shot prompt (asking the model to directly identify a racial covenant); specifically, the \verb|gpt-3.5-turbo-1106| model.\footnote{At the time of the work described in this paper (mid-2024), the two most capable general-purpose LLMs were GPT-4 and GPT-3.5 Turbo, both closed models offered by OpenAI. As GPT-4 would have been much more expensive for the scale of the task, see Section~\ref{sec:resource-cost-comparison}, we excluded it from evaluation.} The input prompt we used can be found in Appendix \ref{sec:instruction-finetuning-prompt}.

Second, we test a few-shot approach with the GPT-3.5 Turbo Model, where two examples of correctly annotated racial covenants are additionally included in the prompt.

Third, we finetune an open LLM for the racial covenant detection task. Given the scale of the racial covenant task, the reliance on an open model poses particular potential benefits: we are able to take advantage of our full labeled training data and the cost of running inference may be substantially lower than paying per API request for GPT. We used Mistral 7B as a base model, then the state-of-the-art open LLM for its size \citep{jiang_mistral_2023}. We finetuned the model with low-rank adaptation (LoRA) \citep{hu_lora_2021} at 16-bit precision on a single A100 80GB GPU. Our custom model was trained on 80\% of our annotations; the remainder (739 pages\footnote{During the evaluation process, we discovered a small number of duplicate pages in our dataset and excluded them from evaluation; 739 pages is therefore slightly less than 20\% of the total dataset.}) was used to evaluate all of the detectors described in this section. The prompt template used for finetuning can be found in Figure~\ref{fig:instruction-finetuning-template}.

For each of our LLM detectors, we also compute a confidence score for the model's classification:
$$\text{confidence} = \text{softmax}(\mathbf{w})_{\text{yes}}$$
where $\mathbf{w} = [w_{\text{yes}}, w_{\text{no}}]$ are the logits for the ``yes'' and ``no'' tokens. We empirically determined that a 75\% confidence threshold resulted in the best performance for the LLM detectors; the evaluation results shown in Section~\ref{sec:evalmetrics} are computed based on that threshold.

\subsection{Racial Covenant Span Recovery} \label{sec:span}

In order to operationally integrate model output for county recorder and counsel review, it was important to pinpoint the racial covenant on the page image, not just in the OCR text. We hence develop an approach to plot a bounding box around the racial covenant provision on a deed page. Our OCR system provides us character-level positional metadata for each page, allowing us to compute a bounding box for any substring of the text. 

In practice, LLMs prompted (or trained) to return an exact span of input text do not always do so, and instead return text with minor variations. This is especially true for text with severe OCR artifacts. To solve this problem, we implement a fuzzy matching algorithm to identify the span of text in a deed page closest to the LLM's output. First, we break the deed text into chunks with a sliding window of size equal to the length of the LLM output. Then, we tokenize each chunk into trigrams and compute the Jaccard similarity between each chunk's tokens and the tokens of the LLM's output. Finally, we select the span with the highest similarity and apply simple heuristics to align the span to sentence and paragraph boundaries.

\subsection{Geolocation}
\label{sec:geo}

One of the major contributions of existing crowdsourcing efforts lies in the geographic characterization of racial covenants. Such geolocation information is important for understanding patterns of housing development, segregation, and mobility, and is hence also important for empirical research on and historical understanding of racial covenants. For instance, testing the long-term effects of covenants on housing segregation requires understanding which properties were and were not subject to racial covenants. 

Unfortunately, the County's pre-1980 microfilm deed records do not contain structured data, such as ownership information, present-day location, or parcel numbers. The initial perception by all parties was that resolving deed records to present day geolocation would be near impossible without significant manual effort. 

\begin{figure}[t]
    \centering
    \includegraphics[width=0.9\linewidth]{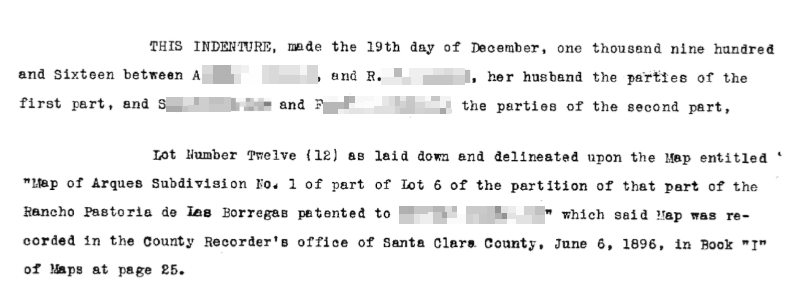}
    \caption{Example of location information in a 1916 property deed. Crucially, we can extract the name of the map which depicts the property, as well as the book and page number on which the map appears. Other useful data, such as the names of the parties and the exact date on which the deed was recorded, can also be extracted.}
    \label{fig:location_in_deed}
\end{figure}

As Figure \ref{fig:location_in_deed} shows, however, the \emph{text} of older deeds does contain some limited information about the location of the properties in question. This text is difficult to parse, referencing, for instance, a ``Map [that] was recorded \dots June 6, 1896, in Book `I' of Maps at page 25.''  

By extracting these textual clues and cross-referencing them against multiple administrative datasets from the County's Surveyor's Office, we show that it is possible to recover the location of individual properties to the tract level (several blocks) for most properties. This is particularly significant as county recorders have engaged surveyors for custom projects to conduct such geo-referencing. 

Our pipeline for geolocating properties consists of several steps:

\begin{enumerate}
    \item \textbf{Information Extraction}: We use a few-shot prompted LLM (\texttt{gpt-4o-mini}) to extract key geographic information from the OCR text, including the name of the referenced map, book number, page number, and other relevant details such as mentioned street names.
    
    \item \textbf{Map Matching}: We cross-reference the extracted map information against a list of maps published by the Santa Clara County Surveyor's Office. Individual maps typically encompass a single developer's subdivision, usually the size of a city block. This step involves fuzzy matching the extracted map name against the names in the spreadsheet to account for OCR errors and variations in naming conventions. The Surveyor's Office records provide exact book and page numbers for each map, as well as scans of the (often hand-drawn) maps themselves.
    
    \item \textbf{Geospatial Lookup}: Using the verified book and page numbers, we query the Surveyor's ArcGIS system to retrieve the precise geographic location of a given map. This system contains geospatial data for many historical maps, allowing us to pinpoint the location of the property in question.
    
    \item \textbf{Manual Augmentation}: For cases where the ArcGIS system lacks coverage, we conducted manual research using scans of historical maps available at the Surveyor's office. We focused on the most frequently occurring maps not found in the digital system, manually determining their present-day locations to enhance our dataset's coverage.
\end{enumerate}

While not exhaustive, this approach allows us to construct a granular picture of the geographic distribution of racial covenants over time, down to at least the level of individual neighborhoods. It also enables us to identify which developers and individuals were instrumental in their proliferation. The resulting dataset allows for a rich analysis of spatial patterns of restrictive covenants and their potential long-term impact on residential segregation in Santa Clara County.

\section{Evaluation}
\label{sec:evaluation}

We evaluate model performance using standard classification metrics on our evaluation set, which comprises 739 pages of human-annotated deeds from Santa Clara County. We also compare this against the resources required to run the model across the full dataset of Santa Clara County deeds. We find that a finetuned open model (Mistral) outperforms all other methods.

\subsection{Evaluation Metrics for Detection}
\label{sec:evalmetrics}

To assess models' abilities to detect racially restrictive covenants, we compute page-level precision, recall, and F1 and span-level BLEU scores~\citep{papineni2002bleu} (on the model's reproduction of the exact span of the racial covenant) on our evaluation set.

These are calculated on an evaluation set of 739 pages of deeds, which were held out and not used for training. Roughly 7 in 10 of the pages in our evaluation set contain a racial covenant. By contrast, our best-performing detector found that fewer than 2 in 1000 deeds in the full 5.2 million deed collection contain one.

AB 1466 requires that county counsel review every provision to be redacted, so high false positive rates (or low precision) can quickly become burdensome.
Meanwhile, a high false negative rate (or low recall) would mean that some racially restrictive covenants are missed by the system.
We use the BLEU score to assess the overlap between the annotated racially restrictive covenant and the identified text. If the BLEU score is low, this may mean that the identified text span doesn't neatly overlap with the annotated text span. Alternatively, it may mean that the language model ``hallucinated'' information that wasn't in the original document~\citep{rohrbach2018object,magesh2024hallucination}.

\subsection{Results for Detection}

\begin{table}[!htbp]
\centering
\begin{tabular}{@{}lllll@{}}
\toprule
\textbf{Model} & \textbf{Precision} & \textbf{Recall} & \textbf{F1} & \textbf{BLEU} \\ \midrule
\textit{Keyword Matching} & 0.913 (0.889, 0.930) & 0.971 (0.955, 0.980) & 0.941 & - \\
\textit{Fuzzy Matching} & 0.992 (0.982, 0.996) & 0.898 (0.873, 0.917) & 0.943 & - \\
\textit{GPT Zero-Shot} & 0.993 (0.982, 0.996) & 0.771 (0.738, 0.799) & 0.868 & 0.787 \\
\textit{GPT Few-Shot} & 0.926 (0.904, 0.942) & 0.961 (0.944, 0.973) & 0.943 & 0.773 \\
\textit{Mistral Finetuned} & \textbf{1.000} (0.995, 1.000) & \textbf{0.994} (0.984, 0.997) & \textbf{0.997} & \textbf{0.932} \\ \bottomrule
\end{tabular}
\vspace{0.5em}
\caption{Page-level precision, recall, F1, and span-level BLEU score for each racial covenant detector, with 95\% Wilson score confidence intervals for precision and recall.}
\label{tab:metrics}
\end{table}

\textbf{Effectiveness of Detection.} Table \ref{tab:metrics} shows a summary of our evaluation.

\begin{figure}[!htbp]
    \centering
    \includegraphics[width=\linewidth]{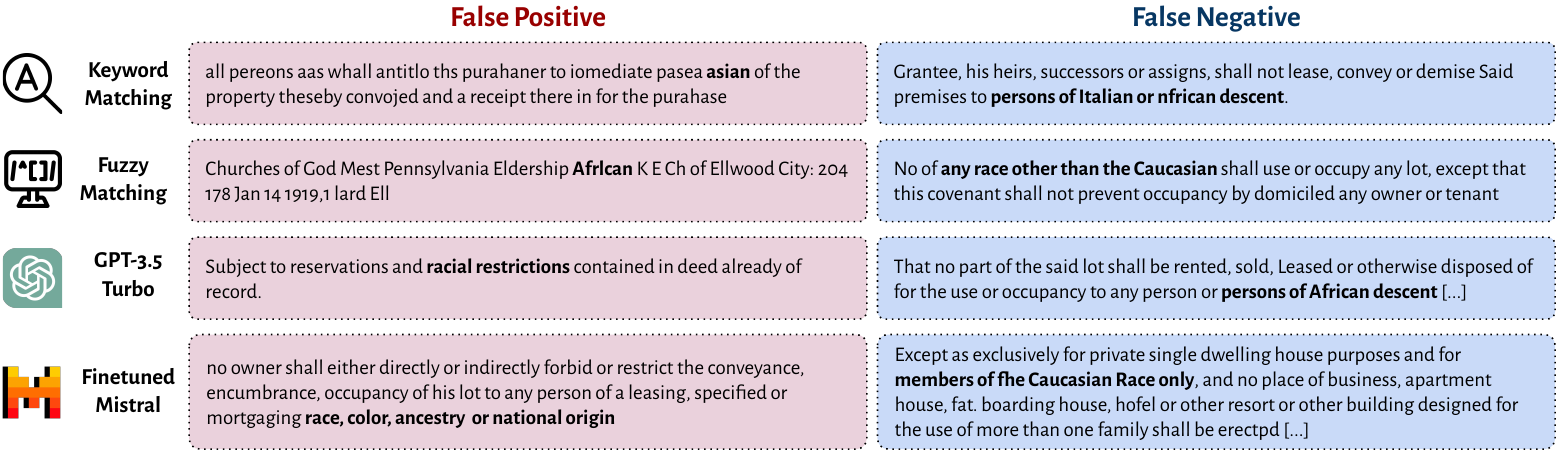}
    \caption{False positive and false negative predictions from each detection approach considered in our study. These examples show typical OCR errors present in our pipeline.}
    \vspace{2mm}
    \label{fig:example_errors}
\end{figure}

First, we find that the finetuned Mistral model uniformly performed better than all the other detectors across all metrics that we examined.
Importantly, the Mistral model is able to identify more racially restrictive covenants than any other method (recall of 99.4\%), while never, in our evaluation set, misidentifying other text as a racially restrictive covenant (precision of 100\%).\footnote{We note in \Cref{ref:integration}, however, that at deployment time we ran additional manual review and identified some false positives in out-of-distribution documents, such as the example shown in Figure~\ref{fig:example_errors}, bottom left.} We also find that the finetuned Mistral model mitigates ``hallucinations'' that are found in other language model-based approaches. 

Second, keyword-based detectors exhibited a range of errors, illustrated in \Cref{fig:example_errors}. For keyword matching, lower OCR quality meant that some racial covenants would be missed due to racial terms being misspelled. Conversely, terms could wrongly appear within a misspelled word, causing a high false positive rate of 28.9\%. The regular expression-based fuzzy matching detector addressed some of these challenges and the false positive rate between the two dropped from 28.9\% to 2.2\%. However, the lack of understanding of context, for both of these approaches, resulted in more significant errors overall.

Third, off-the-shelf language models exhibited only modest improvements, if any, over keyword matching. The initial GPT 3.5 Turbo detector with zero shot prompting had a false negative rate of 22.9\%, likely due to the lack of similar language or tasks in the training data. Few-shot prompting provided the model with more examples of racial covenants. This reduced the false negative rate to 3.9\%, but the false positive rate increased to 23.9\%. The majority of false predictions from the few shot model appeared to be ``hallucinations,'' as evidenced by the lower BLEU score.  

\textbf{Efficiency.} We then compare the costs of the different approaches for processing all 5.2 million Santa Clara County documents, as seen in Table \ref{tab:cost_comparison}.\footnote{Please refer to Appendix~\ref{sec:resource-cost-comparison} for a detailed resource comparison of each approach.} 
Due to the immense volume of property deeds owned by the county, we estimate that manual review, while likely effective, would take 86,667 hours of paid labor for a single individual to complete exhaustively. That kind of review would have required an enormously expensive undertaking, making it nearly impossible to implement AB 1466 in a timely manner. Recall, for instance, that Los Angeles is contracting with a vendor for \$8M to complete the task over a seven year period. 

By contrast, our finetuned Mistral model can process 5.2 million pages for less than \$300 of compute costs on any commodity cloud provider. Consider the cost comparison to closed models. While an off-the-shelf language model like GPT-3.5 Turbo shows promising performance in a few-shot setup, running it across the County's entire collection would cost 43 times as much (more than \$13,000 compared to \$300).\footnote{See Appendix~\ref{sec:resource-cost-comparison}.}

These comparisons illustrate a significant performance and cost-advantage to open models in this context. 

\subsection{Evaluating the Geolocation Pipeline}

Due to the resource-intensive nature of validating our geolocation pipeline, we conduct a smaller-scale sample review. We sample 50 geolocated documents and manually verify correctness. Within this sample, we did not identify any misidentified geolocations, but it is possible that there are errors. Any discussion of specific locations, however, are manually verified.

\section{Integration with Santa Clara County: Processing 5.2 Million Pages of Deeds with AI}
\label{ref:integration}

After identifying the fine-tuned Mistral model as the best option, we developed a pipeline for responsibly processing 5.2 million pages of Santa Clara County deeds. Here, we describe the additional engineering and design efforts for processing this magnitude of documents, including additional manual review by our team and the Santa Clara County officials.

\textbf{Processing 5.2 Million Pages of Deeds.} Following the evaluation pipeline, we scan 5.2 million pages of deeds using our OCR pipeline and apply our model. OCR was the most computationally expensive part of our pipeline, taking 215 hours to complete on 4 A100 GPUs.

We then spot check the deeds with identified RRCs. Unlike our evaluation, we found several false positives in the 1960s and 1970s -- our evaluation dataset has docs from many different time periods outside of Santa Clara, but our Santa Clara County evaluation documents were exclusively pre-1940s.

One example of how the temporal shift led to the emergence of false positives are so-called ``fair housing'' covenants. Several false positives from the 1970s stipulated the opposite of a racial covenant: citing the Fair Housing Act of 1968, these provisions explicitly \emph{banned} discrimination on the basis of race. Because our fine-tuned model was never exposed to any similar language in training, it incorrectly flagged the discussion of racial restrictions as a racial covenant.\footnote{\label{note:fp-example}One such covenant read: ``Fair Housing: No owner shall, either directly or indirectly, forbid or restrict the conveyance, encumbrance, leasing, mortgaging, or occupancy of his unit to any person of a specified race, color, religion, ancestry, or national origin.''}

However, we observed that the model was well-calibrated: more than 90\% of these later-period false positives had a prediction confidence score of below 75\%, while no known true positives did. We therefore set a minimum confidence threshold of 75\% in preparing our final results.\footnote{As an additional heuristic, we also filtered any covenant which contained the phrase ``fair housing'' on the assumption that these were likely to be false positives similar to the example in note~\ref{note:fp-example}.}

To ensure that our results were sufficiently precise, we randomly sampled 200 positive predictions from our final results. We observed two false positives; the precision on the full dataset can be estimated with a 95\% confidence interval of 96.4\% to 99.7\%.\footnote{These figures are calculated using the Wilson score interval for binomial proportions.}

Finally, for any specific covenants that we discuss, we manually review to ensure that identification is correct. And over 4,500 of the identified covenants have already received confirmation through attorney review.

\textbf{Responsible Integration with Santa Clara County Workflow.} Our longer-term partnership enabled working closely together to consider responsible integration of machine learning output. 

\begin{figure}[t]
    \centering
    \includegraphics[width=\linewidth]{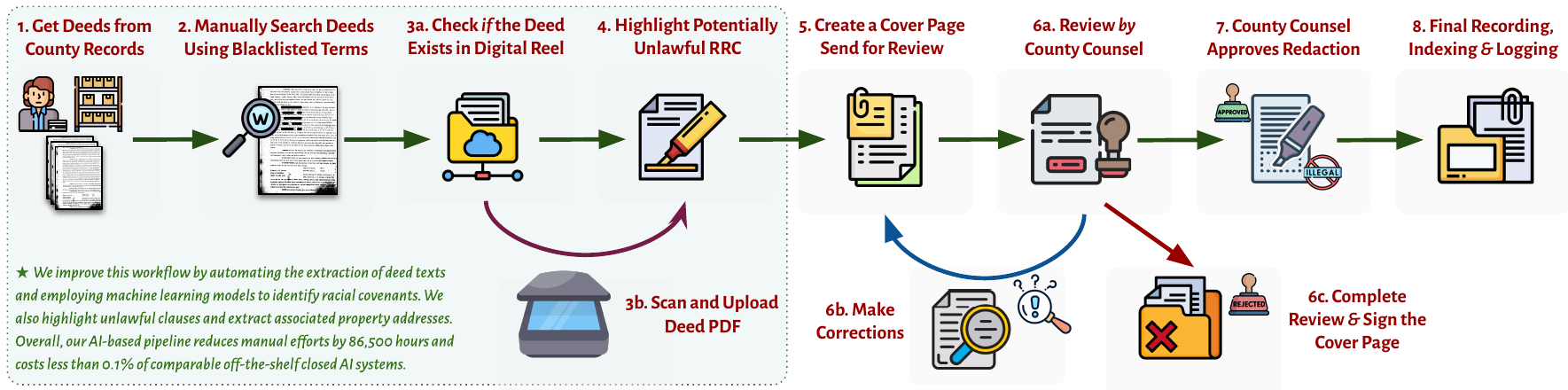}
    \caption{This workflow diagram depicts the multi-step process undertaken by the Restrictive Covenant Modification Program (RECOMP) in Santa Clara County Clerk Recorder’s Office to identify, review, and remove RRCs from real property deeds. The process begins when CRO locates a deed with potential RRCs and checks for a digital copy. If none exists, the original deed is scanned and uploaded to the County Cloud. CRO highlights unlawful language and submits the document to the County Counsel (CC) for review via DocuSign. Based on CC’s review, the redaction may be approved, corrections may be required, or the request may be rejected. If approved, RECOMP proceeds with the redaction, records, indexes, and verifies the document. The final version is then logged and uploaded for accurate and up-to-date recordkeeping.
}
    \label{fig:integration}
\end{figure}

Figure~\ref{fig:integration} presents the baseline operational workflow for how the County Recorder and the County Counsel offices approached the Restrictive Covenant Modification Program (RECOMP). Our model essentially took the time-consuming search steps 1--3 and automated them. After that, the county recorder reviewed the results and delivered them to the County Counsel for approval of redaction, per statutory requirement under AB 1466. All redacted provisions were retained, as required.  

Government integration of technology -- and AI specifically -- has been challenging \citep{pahlka2023recoding}.  Procuring AI systems or technology from private vendors can be particularly difficult  given the changing nature of technology and contracting process \citep{kelman1990procurement}. On the other hand, civic technology can generate many ideas and prototypes, but public agencies require long-term engagement to integrate, monitor, and evaluate the benefits of technology \citep{engstrom_government_2020}.
Our partnership enabled us to identify the most time-consuming task (search) and test the benefits of AI assistance. The academic team ensured the evaluation of performance, and the County team reconfigured its operational processes to integrate the model output. One key principle is that AI is used solely in aid of the statutorily required human review, ensuring that any redactions meet the county counsel standard for an RRC. 

Put differently, rather than promising an end-to-end solution, our partnership enabled us to identify the biggest pain point that could be addressed by AI, while keeping humans responsibly in the loop. 

\section{The Evolution of Racial Covenants in Santa Clara County} \label{sec:evolution}

Our analysis of Santa Clara County deeds offers a detailed examination of racial covenants over space and time. As we note in \Cref{sec:evaluation}, our final review sample contained a small amount of false positives and we suspect that there may be a small number of additional RRCs of unusual construction below the 75\% confidence threshold. Taking into account the relatively small error rate, we characterize some trends that we identified. 
From 1907 to 1974, we identified  roughly 7,500 deeds that the County recorded containing racial covenants. We first present results on the geographic distribution of these covenants (\S~\ref{subsec:map}). Second, we show how our data enables us to identify the small number of developers disproportionately responsible for racial covenants (\S~\ref{subsec:developers}). Third, we 
provide a characterization of the historical evolution of racial covenants that are distinctly informed by our new Santa Clara County data (\S~\ref{subsec:periods}). Overall, our findings illustrate how machine learning can support the implementation of redactions, while unearthing historical discrimination in a more granular fashion. 

\subsection{Geographic Distribution}
\label{subsec:map}

\begin{figure}
    \centering
    \includegraphics[width=\linewidth]{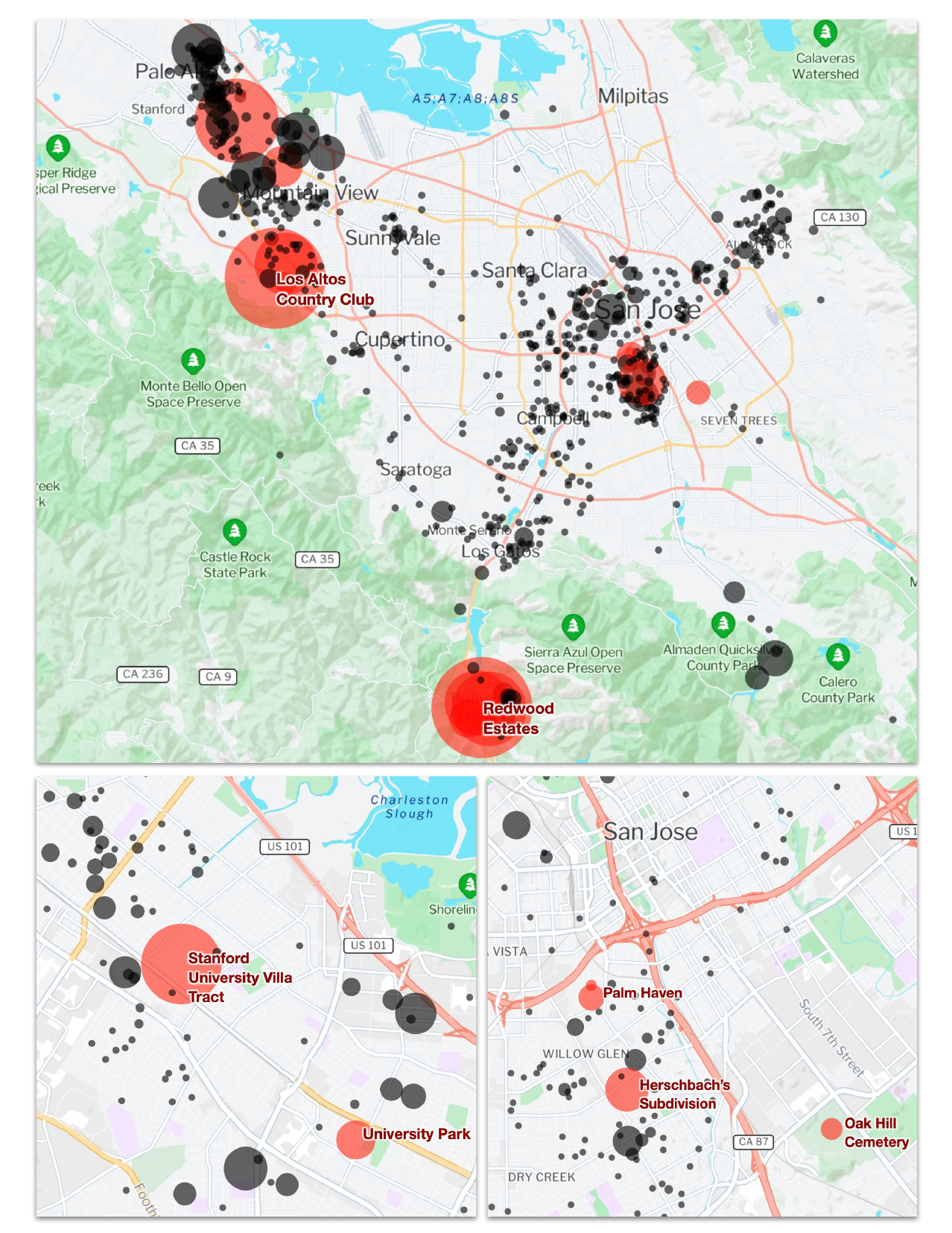}
    \caption{\textbf{Top:} Clusters of racial covenants on a map of modern-day Santa Clara County. Some of the largest and most notable racially restricted developments -- discussed in this section -- are shown in red. \textbf{Bottom left:} Racial covenants in south Palo Alto and Mountain View. \textbf{Bottom right:} Racial covenants in downtown San Jose. Dots represent individual subdivisions and are scaled in proportion to the number of racial covenants within the subdivision.}
    \label{fig:scc_rich_map}
\end{figure}

\begin{figure}[ht]
    \centering
    \includegraphics[width=\linewidth]{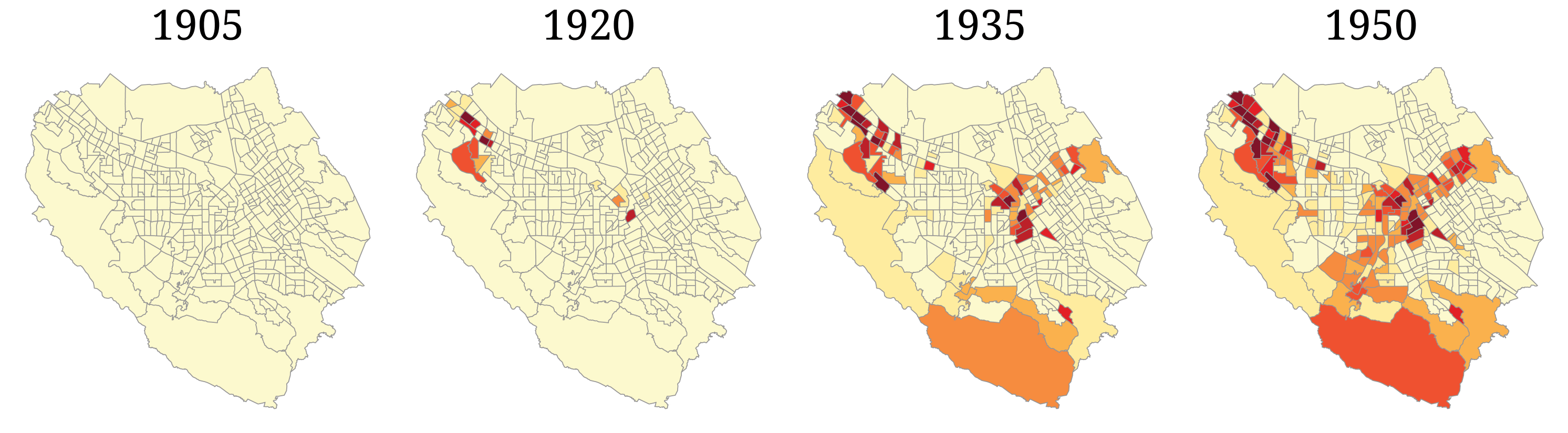}
    \caption[]{Density of properties with racial covenants in modern-day Census tracts in 1905, 1920, 1935, and 1950. Racial covenants are plotted cumulatively. Racial covenants are initially concentrated in modern-day Palo Alto and San Jose, but spread throughout the county between 1920 and 1950. Some tracts in the south and east of the County are omitted here for space reasons; a plot of the full County can be found in Appendix~\ref{appendix:full_faceted_map}.\vspace{0.5em}}
    \label{fig:scc_map_faceted}
\end{figure}

Our geolocation pipeline, described in \S~\ref{sec:geo}, is able to identify the tract-level locations of 79.0\% of properties with racial covenants. Figure \ref{fig:scc_rich_map} plots the locations of properties subject to racial covenants, with dots proportional to the number of covenants in each tract. Figure \ref{fig:scc_map_faceted} depicts the spread of racial covenants throughout the County over the first half of 20th century.

Prior to 1920, racial covenants were a highly concentrated phenomenon in Santa Clara County, driven by a small number of large developers. In this early period, the overwhelming majority of covenants were attached to properties located in modern day Palo Alto and San Jose. The two largest racially restricted developments in the County at this time were the University Park and Stanford University Villa tracts, both developed with the creation of the Stanford university (lower left panel of Figure~\ref{fig:scc_rich_map}).\footnote{See, e.g., Stanford University: Some of the Building to be Opened for the Next Fall Term, \emph{The Evening News}, Feb.\ 8, 1889, at 3.}

\begin{figure}
    \centering
    \includegraphics[width=\linewidth]{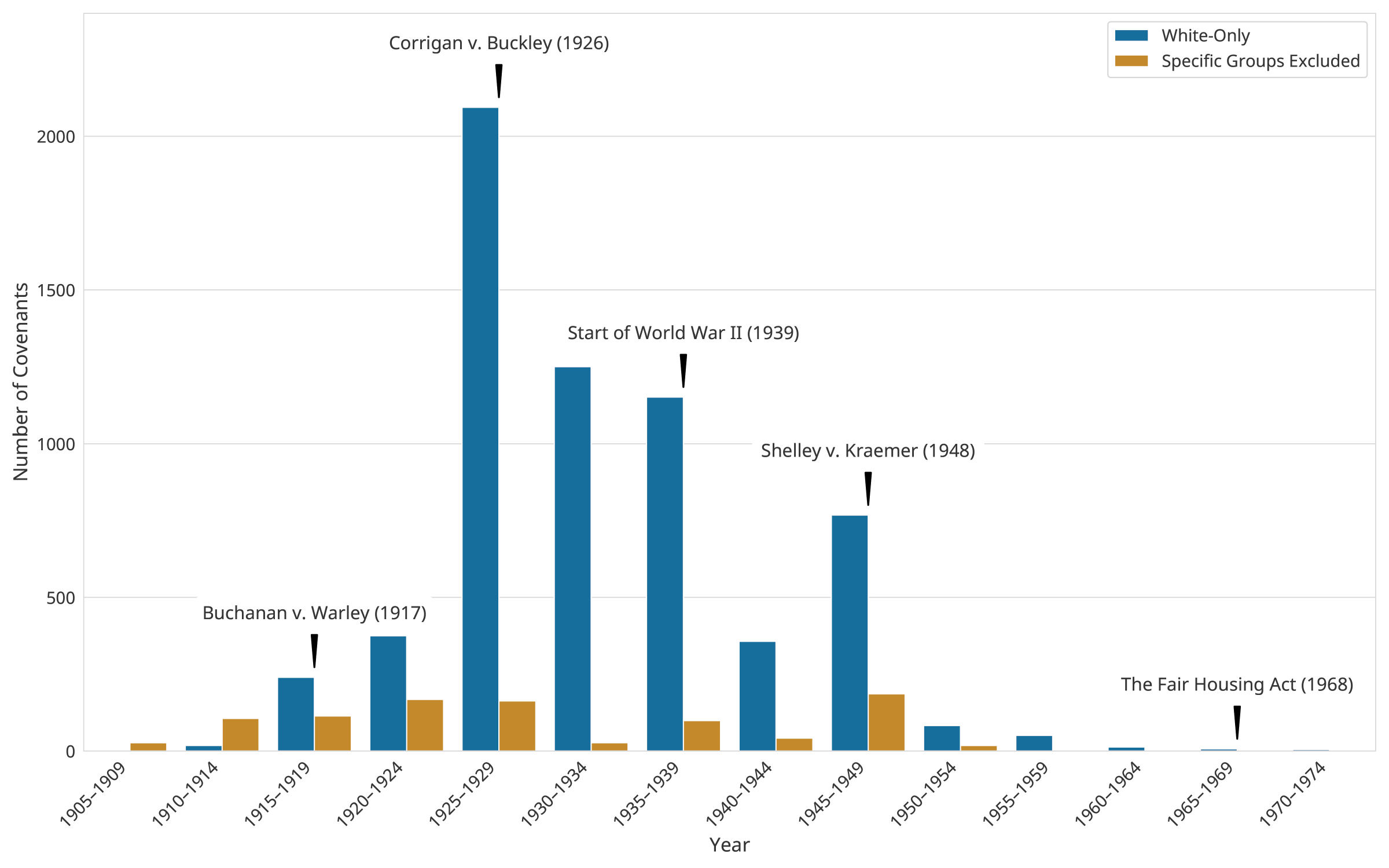}
    \includegraphics[width=\linewidth]{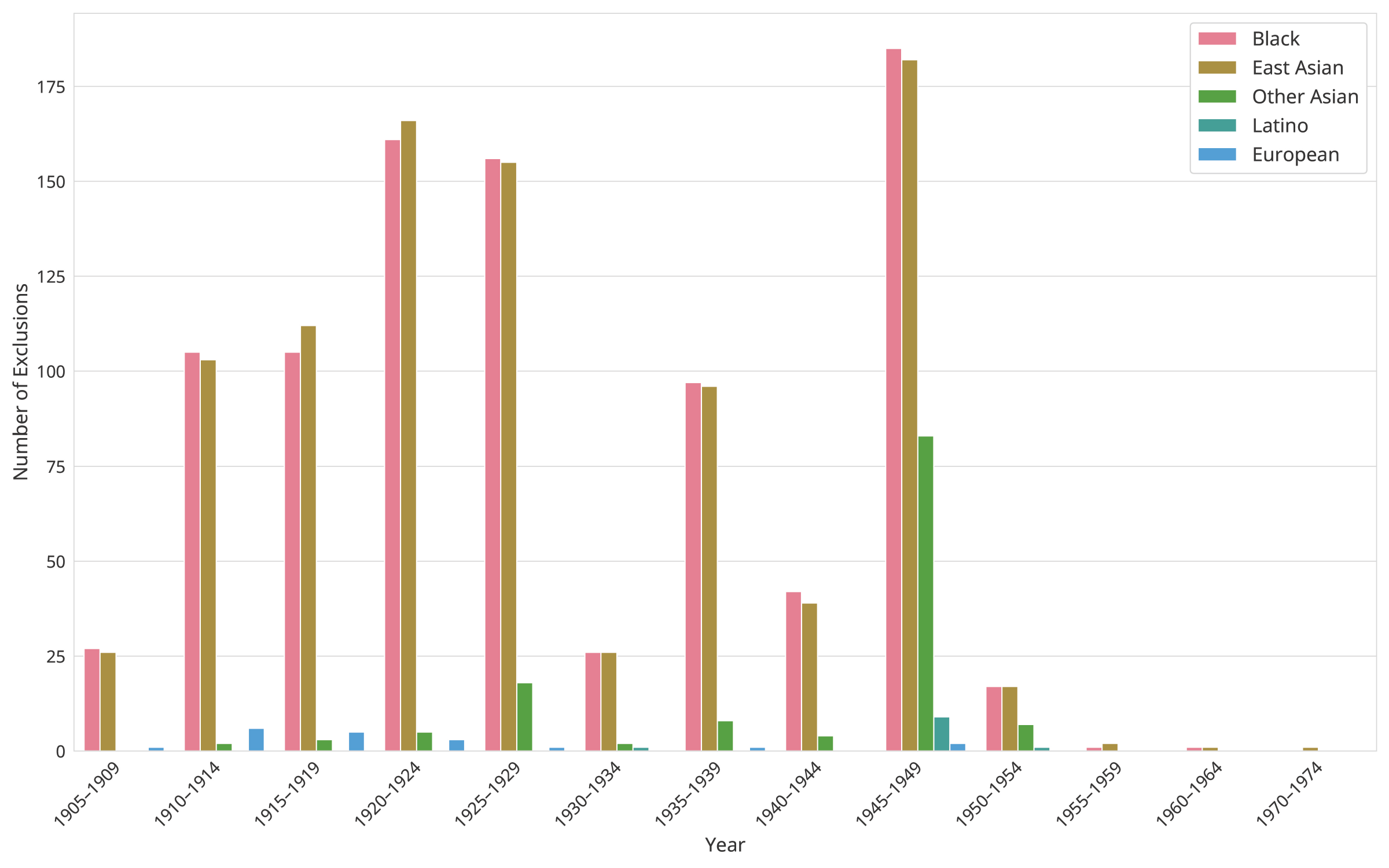}
    \caption[]{\textbf{Top:} Number of property deeds with restrictive covenants from 1905--1974, divided by whether specific racial groups were excluded or only white/Caucasian individuals were permitted. Most pre-1915 covenants specifically exclude Black and Asian individuals, but the vast majority of later covenants are white-only. The small number of restrictive covenants matched after 1970 consists largely of older deeds filed for reference, rather than new restrictive covenants being introduced. \textbf{Bottom:} The number of occurrences of specific racial groups in covenants that exclude specific groups. East Asian and Black were by far the most commonly excluded demographics, but some covenants targeted other groups, such as Italian, Portuguese, Indian, and Mexican individuals.}
    \label{fig:covenant-distributions}
\end{figure}

As Figures \ref{fig:scc_map_faceted} and \ref{fig:covenant-distributions} show, racial covenants became dramatically more common across the County during the 1920s and 1930s. During this period, racially restricted properties became more diffuse, and it became increasingly common for individual sellers to insert discriminatory provisions on sale of their property. 1925 also saw the first use of a racial covenant at San Jose's Oak Hill Cemetery -- then publicly owned by the City of San Jose~\citep{rhoads_cemetery} -- which would go on to sell at least 50 burial plots for the sole use of ``the human dead of the Caucasian race'' (lower right panel of Figure~\ref{fig:scc_rich_map}).\footnote{In 1933, the Oak Hill Cemetery was sold to a private party, though it continued to sell racially restricted burial plots for more than a decade thereafter. The cemetery, now known as the Oak Hill Memorial Park, continues to operate to this day~\citep{rhoads_cemetery}.} The fact that the city of San Jose owned the cemetery, with racial covenants on burial deeds, complicates conventional historical accounts of racial covenants as marking the shift from discrimination by state action to by private action.\footnote{Since the late 1800s, the city had chartered and contracted with the Oak Hill Improvement Company and the Oak Hill Cemetery Association to operate the cemetery. Burial deeds were hence sold from the Association, while the city retained control of the land until the 1933 sale. See Cemetery Affairs: City May Sue Oak Hill Improvement, \emph{The Evening News},  Jan.\ 22, 1900, at 1. This anticipates questions of when a private business is sufficiently connected to government to be deemed a state actor. See, e.g., Burton v.\ Wilmington Parking Authority, 365 U.S.\ 715 (1961). Notably, in 1900, several Chinese associations purchased land adjacent to Oak Hill, where Chinese were not allowed to be buried, to create a Chinese Cemetery. See \url{https://thebpog.org/chinese-american-cemetery/}.} 

Notably, a rural community in the Santa Cruz Mountains known as Redwood Estates accounts for 796 covenants, more than 10\% of all found in the County. With a 1940 population of less than 4,200 individuals, it is likely that the entire town was covered by racial restrictions.\footnote{According to the 1940 decennial Census, the population of ``Redwood Township'', the enumeration district which included Redwood Estates, was 7,822. Excluding the town of Los Gatos, the remainder of the township's population was 4,225. While a population estimate for Redwood Estates alone is not available, several other rural communities existed within the township (including 96 racial covenants filed outside Redwood Estates), implying an upper bound of less than 4,200 people \citep{us_census_bureau_1940_nodate}.}

After 1940, we find that use of racial covenants steadily declined, first with the onset of World War II (potentially due to a  decline in new construction) and later with \emph{Shelley}. However, racial covenants did continue to proliferate in smaller numbers throughout the County, especially in areas of new development such as the city of Mountain View.

\subsection{The Role of Developers}
\label{subsec:developers}

Our analysis also shows that a small number of developers appear responsible for the vast majority of racial covenants. We calculate that merely ten developers appear responsible for roughly a third of covenants.  Part of the explanation here lies in the growth of the Santa Clara County population during this period, nearly tripling in population from 1920--1950. Developers played a prominent role in the construction of Santa Clara County (and, in turn, Silicon Valley), converting agricultural land into residential neighborhoods. 

\begin{figure}[pt]
    \centering
    \includegraphics[width=3.75in]{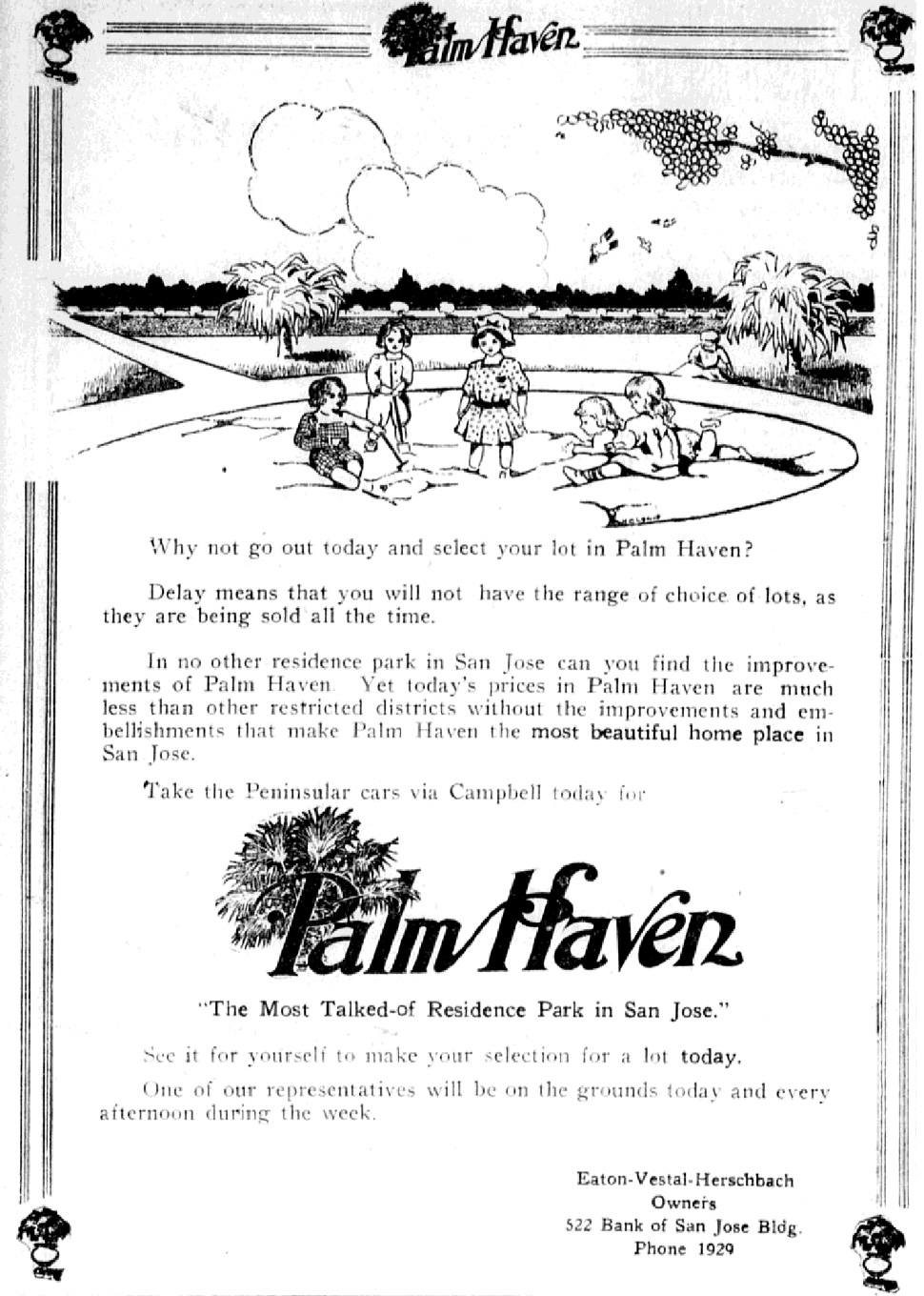}
    \caption[]{1913 housing advertisement for the Palm Haven neighborhood in San Jose. 1913 is several years before \emph{Buchanan} found ``restricted districts'' based on race unconstitutional, and the advertisement emphasizes the ``restricted district[].'' Palm Haven construction dates straddled \emph{Buchanan}. It was developed by Thomas Herschbach who came to be responsible for 161 racial covenants in the County.}
    \label{fig:palmhaven}
\end{figure}

Thomas A. Herschbach, for instance, was a prominent developer and builder of numerous subdivisions around San Jose, described laudably in historical volumes \citep{halberstadt}. When the Stone Church of Willow Glen Presbyterian ran out of funds, Herschbach, the builder, donated the roof (id.).  Yet not told in these volumes is the fact that Herschbach was single-handedly responsible for 161 deed records with racial covenants. In the new Palm Haven subdivision that he developed in the years before \emph{Buchanan},  advertisements bearing his name described the development as ``the most beautiful homeplace in San Jose,'' with a picture of six evidently white children playing in a sandbox, and touting a price lower than ``other restricted districts'' (Figure~\ref{fig:palmhaven}).\footnote{\emph{Sunday Mercury and Herald}, May 4, 1913, at 8.}  In the years of development, \emph{Buchanan} would strike down such restrictive racial zoning, and hence Herschbach developments turned to racial covenants.  Scores of local newspaper advertisements continue to advertise the ``restricted'' nature of properties in Santa Clara County post-\emph{Buchanan}, marking the linguistic shift from government-sanctioned to privately-enabled ``restrictions.'' As one property advertisement in Willow Glen, a neighborhood Herschbach developed, put it in 1946: ``Willow Glen's finest and most attractive subdivision \dots High building and \emph{racial restrictions}.''\footnote{\emph{San Jose Mercury Herald}, June 8, 1946, at 11 (emphasis added).}

Herschbach was far from alone, and we found that several other prominent members of California society were prolific spreaders of racial covenants. For example, Virginia M.\ Spinks, who served as an elector for Woodrow Wilson in the 1916 presidential election\footnote{Complete Totals General Election, \emph{San Jose Mercury Herald}, Nov.\ 21, 1916.} and later in his administration\footnote{Executive Order 2745, Authorizing Appointment of Virginia M.\ Spinks to Position in Department of Labor Without Regard to Civil Service Rules, Nov.\ 1, 1917.}, sold at least 87 properties with racial restrictions attached. The disproportionate role of a small number of sellers also challenges the notion that racial covenants were primarily a signaling device of more ``loosely knit'' communities, at least in California~\citep{brooksrose2013saving}.\footnote{Our evidence is not necessarily inconsistent, but the setting in Santa Clara County raises the question of how to define whether a community is ``loose-knit.'' At the relevant historical period, the County is (a) relatively homogeneous (with very few non-white residents), and (b) covenants are imposed by a small number of developers for new communities. The relevant conception may be about the homogeneity of potential purchasers, which is difficult to measure.}

Our findings also point to the critical role of \emph{agency} exercised by and responsibility of individual developers, a small number of whom developed large subdivisions that converted agricultural to residential land. 
Historical accounts on this diverge, with conventional narratives noting that ``crusaders'' for racial equality would lose business \citep{taeuber1961privately}. Others, however, have pointed to the role of Joseph Eichler, who developed many Palo Alto divisions and refused to adopt racial covenants. Eichler aimed to quietly demonstrate that integration could be good business, and came to exert influence on fair housing policy \citep{howell}.  Our evidence suggests that in an emerging housing market, where a small number of developers, such as Eichler and Herschbach, were market movers, and where racial covenants were still in the minority of overall deed records post-\emph{Buchanan}, agency may have been possible \citep{redford}.

\subsection{Historical Evolution}
\label{subsec:periods}

\paragraph{1. Emergence (1907--1916).}

The earliest instances of racial covenants date to 1907, with explicit language excluding multiple ethnic groups: ``the party of the second part also agrees for himself, his heirs, executors, administrators, or assigns that he will not sell; nor lease nor permit the premises to be occupied by Italians, Portuguese, colored people or Spaniards.'' Covenants codified racial hierarchies into private law. 

This early period coincided with the rise of nativist movements across the United States, fueled by fears that non-white and non-Anglo-Saxon groups would ``dilute'' white communities. In Santa Clara County, these anxieties were reflected in demographic shifts, as seen in Table~\ref{table:scc_population}, where the Black population declined from 989 in 1890 to 262 in 1910, and the Chinese population fell from 2,723 to 1,064. At the same time, the Japanese population saw a sharp increase, rising from just 27 in 1890 to 2,299 by 1910. These population dynamics likely amplified the perceived need among white property owners to impose restrictions on land ownership and occupancy.

The language of these early covenants reveals the racial hierarchies prevalent at the time, with African Americans, Asians, Latinos, and Mediterranean immigrants often singled out as ``undesirable.'' These groups were considered culturally incompatible with the aspirations of white, middle-class neighborhoods, which sought to maintain racial and cultural homogeneity. 

\begin{table}[!t]
\centering
{\renewcommand{\arraystretch}{1.1}
\scalebox{0.95}{
\begin{tabular}{l  r r r r r r r}
\toprule
 & \textbf{Total} & & & \textbf{Native} \\
\textbf{Year}  & \textbf{Population} & \textbf{White}     & \textbf{Black}    & \textbf{American}  & \textbf{Chinese}  & \textbf{Japanese} & \textbf{Others} \\
\midrule
1890  & 48,005  & 44,247  & 989    & 19     & 2,723   & 27      & 0      \\
1900  & 60,216  & 57,934  & 251    & 9      & 1,738   & 284     & 0      \\
1910  & 83,539  & 79,849  & 262    & 16     & 1,064   & 2,299   & 49     \\
1920  & 100,676 & 96,471  & 335    & 4      & 839     & 2,981   & 46     \\
1930  & 145,118 & 138,589 & 536    & 45     & 761     & 4,320   & 867    \\
1940  & 174,949 & 168,921 & 730    & 74     & 555     & 4,049   & 620    \\
1950  & 290,547 & 280,429 & 1,718  & 144    & 685     & 5,986   & 1,585  \\
1960  & 642,315 & 621,625 & 4,187  & 705    & 2,394   & 10,432  & 2,972  \\
1970  & 1,064,714 & 1,003,898 & 18,090 & 4,048  & 7,817   & 16,644  & 14,217 \\
1980  & 1,295,071 & 1,030,659 & 42,835 & 10,011 & 22,745  & 22,262  & 166,559 \\
\bottomrule
\end{tabular}
}
}
\vspace{0.5em}
\caption{Racial demographics of Santa Clara County from 1890 to 1980, based on U.S. Decennial Census data. The table shows population of various racial and ethnic groups over the period, reflecting significant changes in the county's racial composition as the population grew from approximately 48,000 in 1890 to nearly 1.3 million by 1980. The white population dominated the total population, increasing from 44,247 in 1890 to over 1 million in 1980, though its share of the total population declined as the County became more diverse. The African American population saw substantial growth, particularly after 1950, expanding from 989 in 1890 to over 42,000 by 1980. The Native American -- originally referred to as Indian in the Census Data -- population remained small but increased from just 19 individuals in 1890 to over 10,000 in 1980. Significant growth is observed in Asian populations, particularly among the Chinese and Japanese communities. By 1980, the Chinese population had grown to 22,745, while the Japanese reached 22,262. The ``Others'' category, which includes individuals not listed in the defined racial groups, shows considerable growth. The spike in 1980 likely reflects changes in the Census treatment of race as distinct from ethnicity (Hispanic), following a 1977 OMB directive.}
\label{table:scc_population}
\end{table}

\paragraph{2. Post-\emph{Buchanan} Growth (1917--1926).} The use of racial covenants expanded significantly following \emph{Buchanan}. The case shifted state racial zoning into private restrictive covenants -- seen as out of reach of the Fourteenth Amendment for lack of state action -- as an alternative means to maintain racial boundaries in housing. 

In Santa Clara County, the frequency of racial covenants surged during this period, with annual occurrences rising from 62 in 1917 to over 400 by 1926. This six-fold increase reflects the broader national trend of private actors taking on the role of enforcers of segregation. The language of the covenants also became more specific, particularly targeting African Americans, Japanese, Chinese, and other non-white groups. This shift aligned with the growing anti-immigrant sentiment that culminated in the passage of the Immigration Act of 1924, which severely restricted immigration from Asia.

This  period also coincided with the first Great Migration, leading some to describe the use of racial covenants as ``Jim Crow of the North.''\footnote{PBS, Minnesota Experience: Jim Crow of the North, \url{https://www.pbs.org/video/jim-crow-of-the-north-stijws/}}  So too, we find, in the West. Covenants from this era often included broad racial terms such as ``Mongolian'' and ``Negro,'' as well as references to specific ethnic groups like the Japanese and Chinese, who were viewed as economic competitors in industries like agriculture.

The rise in racial covenants also reflect a growing formalization of discriminatory practices within the real estate industry. Real estate boards, developers, and homeowners increasingly viewed covenants  as essential tools for protecting property values and maintaining racial homogeneity. This was particularly true in suburban developments, where new housing tracts were often marketed as ``restricted'' communities, promising potential white buyers that racial minorities such as African and Asian Americans would be barred from purchasing homes.  

\paragraph{3. Peak Period  (1927--1938).} The 1927--1938 period represents the height of racial covenant usage in the County, spurred by \emph{Corrigan v.\ Buckley}, which legitimized the use of racial covenants as private contractual agreements lacking state action required for constitutional coverage. White property developers, homeowners, and real estate boards escalated their reliance on these covenants as a primary tool for maintaining racially homogeneous neighborhoods. In 1928, Santa Clara County recorded over 600 covenants.  

The language in these covenants became more targeted and explicit. Deed records reveal widespread exclusion of specific ethnic groups, including African Americans, Chinese, Japanese, and other non-Caucasian communities. Terms such as ``Negro,'' ``Mongolian,'' and ``colored'' were commonly employed to delineate the racial boundaries of acceptable property owners and tenants. 

The rise of racial covenants during the late 1920s and early 1930s must also be viewed within the broader context of the post-World War I social climate and the Great Migration. The arrival of African Americans and other minority groups in northern and western cities created heightened racial anxieties among white homeowners, who sought to safeguard their communities through these legally sanctioned racial barriers. What is striking is that the rate at which racial covenants explicitly excluded Black purchasers was at the same  rate as that of Asian exclusion, despite much lower presence of Black residents in the county and the Chinese exclusionary period. This pattern is particularly striking, and corroborates other historical accounts that note that integrating a tract with Asian Americans was a ``minor issue,'' but that selling to African Americans was ``much more controversial and potentially damaging'' \citep{howell}.  As Section~\ref{subsec:developers} illustrated, restrictions via racial covenants were also increasingly marketed as a method of preserving property values, with the implicit understanding that racial segregation would prevent economic decline in white neighborhoods. This perceived linkage between racial homogeneity and property value preservation was deeply embedded in the logic of racial covenants during this era.

The Federal Housing Administration (FHA), known more broadly for promoting redlining, actively promoted the use of racial covenants as a condition for insuring mortgages. The FHA’s underwriting policies explicitly tied the stability of neighborhoods to racial homogeneity, further embedding the use of covenants in the development of new housing projects.

The FHA’s endorsement of racial covenants gave them an air of official legitimacy, encouraging real estate developers to incorporate these restrictions into the blueprints of suburban developments across the country, including in Santa Clara County. During this period, racial covenants not only persisted in existing neighborhoods but also proliferated in new developments as the housing market began to recover from the Depression. The covenants from this era reflect a deepening of racial and socioeconomic exclusions, as terms targeting minority groups became more codified and entrenched within the fabric of real estate transactions.

\paragraph{4. World War II Era Fluctuations (1939--1947).}

World War II brought about significant changes in racial dynamics. The wartime demand for labor, combined with executive orders promoting fair employment practices, allowed African Americans to gain greater access to jobs in defense industries and other sectors.  At the same time, the internment of Japanese Americans forcibly uprooted Japanese American communities across the West Coast. 

The post-war period saw a sharp resurgence of racial covenants. The return of soldiers and the end of wartime economic controls created a surge in demand for housing, and white homeowners and developers once again turned to racial covenants as a means of protecting their neighborhoods from racial integration.

\paragraph{5. Post-Shelley Decline (1948--1967).} \emph{Shelley} was a pivotal moment for racial covenants, as is reflected in Santa Clara County records. Racial covenants significantly and nearly immediately drop, with a near 75\% decrease in prevalence. %
While \emph{Shelley} made racial covenants unenforceable, \citet{brooksrose2013saving} argue that such covenants continued to play an important signaling role. Consistent with their account, homeowners and developers in Santa Clara County continued to include such covenants through the 1950s and 1960s. 

The Fair Housing Act (FHA) of 1968 prohibited racial discrimination in the sale, rental, and financing of housing, making racial covenants illegal, not just unenforceable.

\section{Prevalence}
\label{sec:prevalence}

\begin{figure}[pt]
    \centering
    \includegraphics[width=4.25in]{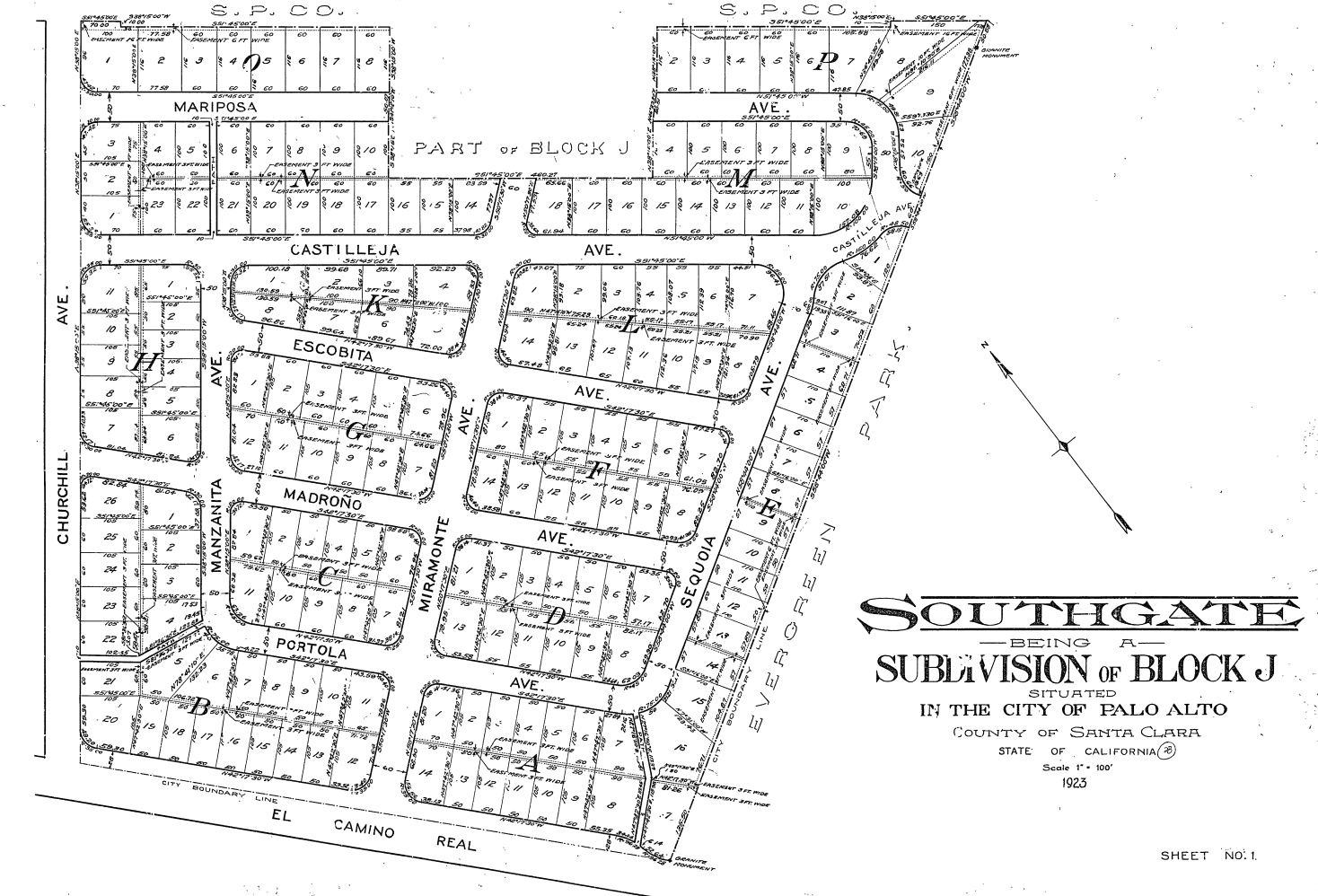}
    \caption[]{Assessor's map for Southgate neighborhood in Palo Alto from 1923. The entire neighborhood of then-196 homes is covered by one racial covenant from the Palo Alto Development Company, which would go on to sell individual properties.}
    \label{fig:southgate}
\end{figure}

So far, we have focused on deeds as the unit of analysis. A single deed record, however, may apply to multiple units, including tracts and neighborhoods. Focusing on the deed record makes sense for implementation under AB 1466, as only a single deed record needs to go through the redaction process. But the raw count of 7,500 deeds may significantly underestimate the affected number of properties. The single covenant document for Palo Alto's Southgate neighborhood, for instance, covers the entire subdivision of 196 homes depicted in Figure~\ref{fig:southgate}.

To estimate the total number of properties covered by racial covenants, we design the following workflow. First, with a combination of keyword heuristics and few-shot LLM classification, we identify pre-1950 deed covenants which appear to cover entire tracts / neighborhoods.\footnote{To provide some intuition on this detection, neighborhood covenants are often referred to as ``Declarations,'' as opposed to ``Deed Records.''}
We find 412 such neighborhood covenants, each of which reflects a large tract being subdivided for sale to individuals.\footnote{Because of their disproportionate impact on our estimates, we manually confirmed that all of these matches contained racial covenants and applied to an entire tract.} Second, we employ the same map matching strategy described in \S~\ref{sec:geo} to identify surveyor's maps representing those tracts. Consistent with our results above, we are able to automatically match and geolocate 79\% of neighborhood covenants to specific maps. Third, for the 86 unmatched deeds, we review the deed record to identify the corresponding surveyor's map (and associated geographic location). Fourth, we inspect the original scans of each map to count the number of lots, reflecting individual housing units, in the tract. This yields a total of 18,871 lots covered by only 412 neighborhood-wide covenants.

Next, we use a language model (\texttt{gpt-4o-mini}) to identify pre-1950 deeds that apply to more than one lot, but do not cover an entire tract.\footnote{The model extracted out a list of lots mentioned for each deed. We verified the accuracy of this labeling with a random sample of twenty units.} This adds another 5,354 individual lots, associated with 1,293 deeds. An additional 5,612 covenants apply to only a single lot.

Last, we deduplicate our counts on a property level to isolate the number of affected properties. Multiple deed records for the same property may reflect sales of the same property, but our aim is to convert deeds into properties affected, regardless of the number of sale events during our observation period. To ensure that the same property is not double-counted, we deduplicate on the combination of (tract, block number, lot number) and filter individual covenants filed within tracts already covered by a community-wide covenant. This process removes 5,315 lots from our count.

Overall, we then have a count of 24,522 lots that were subject to racial covenants. 

To understand the relative magnitude of these restrictions, we use the Decennial Census, which report a total number of 56,406 dwelling units in 1940 and 92,315 in 1950.\footnote{\url{https://www2.census.gov/library/publications/decennial/1950/hc-1/hc-1-48.pdf}} This provides us with the following estimate: in 1950, \emph{one in four} properties were covered by a racial covenant. In 1940, nearly 30\% of all properties were covered by a racial covenant.\footnote{This is based on a count of 16,553 pre-1940 deeds.} This confirms that the identification of 7,500 deeds vastly understates the true impact of racial covenants in Santa Clara County. Because the county tripled in population from 1920 to 1950 (see Table~\ref{table:scc_population}) and the housing stock doubled from 1940 to 1950 -- at precisely the peak usage of racial covenants -- vast portions of the county were covered by racial covenants.  

We note that this estimate is based on a set of assumptions. It provides the first population-level estimate of what fraction of the housing stock was encumbered by racial covenants, and it is subject to uncertainty in resolving deeds into properties.  Factors that may bias this estimate downward are: (a) failure to detect neighborhood-wide covenant language; and (b) subdivision of lots into multiple dwelling units (e.g., into apartments). Factors that may bias this estimate upward are: (a) improper deduplication, (b) failure to develop a subdivided lot, and (c) the fact that a small percentage of covenants were written to expire after some number of decades.  Other factors that have unclear effects on the magnitude are potential inaccuracies in our map and deed matching process. Our best assessment is that the factors leading to undercounting predominate, making the estimate that one in four properties is affected closer to a lower bound. 

The bottom line, however, is simple: racial covenants were pervasively used across Santa Clara County.\footnote{Appendix~\ref{appendix:lot-coverage} provides several visualizations of the lot-level coverage of racial covenants in the County.}

\section{Limitations}
\label{sec:limits}

We note several potential limitations to our approach. First, the use of machine learning may be seen to remove the intrinsic value of volunteer-based efforts that require close engagement with historical sources.  We acknowledge this trade-off, but we also note that the scale of deed records can make exclusively volunteer-based approaches impossible as a solution to documenting racial covenants across the country.  Moreover, we note that there are promising ways for citizens to engage with AI systems and model outputs -- such as through the interactive map that we provide -- that do not require them to function as a labeling workforce \citep{gray2019ghost}. Without needing volunteers to scale manual review, AI assistance can hence make room for distinct types of community engagement, such as around exploration of results, connection with other historical sources, and reflections on implications. 

Second, while our performance evaluation demonstrates remarkably strong results, wherein the AI system appears to spot racial covenants even in the face of serious OCR errors that can fool humans (see, e.g., Figure~\ref{fig:ocr-challenges}), our system does not have perfect recall. It may miss some racial covenants. It is, however, not clear that, relative to near perfect precision and 99.4\% recall, human performance would be any better. Human annotators can differ in quality, get tired after reading many documents, and miss the proverbial needle in the haystack.  

Third, our model may miss some covenants that are more subtle in nature. As \citet{rose2023general} notes, in \emph{Schulte v.\ Starks},\footnote{213 N.W.\ 102 (Mich.\ 1927).} the Michigan Supreme Court upheld the interpretation of a Detroit covenant that prohibited purchasers who ``would be injurious to the locality'' -- without explicit mention of race -- as barring African Americans. Humans may, of course, also miss these, and AI offers us a chance to prompt (via few-shot learning~\citep{wang2020generalizing}) for such boundary covenants. 

Fourth, and related, California civil rights law prohibits discrimination not just on the basis of race, but on the basis of many other protected attributes, including age, religion, sex, gender identity, familial status, disability, veteran or military status, national origin, genetic information, or source of income. We focused here on the predominant concern animating AB 1466, namely racial covenants, but county recorders may also need to assess for the presence of other restrictive covenants. While we do not explore that here, our general approach facilitates the discovery of such rare provisions through few-shot learning or further fine-tuning \citep{wang2020generalizing}. Indeed, our research has already uncovered instances of covenants plausibly based on family status, showing the potential path for a broader sweep to surface non-racial covenants.\footnote{For instance, one covenant indicates that any building is ``to be occupied by only one family.''} %

Last, some might object to the efforts to systematically redact racial covenants. Resources could be put to better use for affirmative housing reform, as these covenants cannot be enforced. Other commentators favor putting the onus on homeowners to redact deed records. Carol Rose, for instance, takes issue with proposals to redact deed records en masse, stating,  ``If these things are taken out of the record books, [they're] gone. It’s like Stalin pushing a button and saying ‘delete Bukharin.’''\footnote{Carol M.\ Rose, ``De-racing Property: Earl Dickerson and the Struggle Against Racially Restrictive Covenants,'' Dickerson Conference: Business Person and Movement Lawyer, University of Chicago Law School, Oct.\ 30, 2020, \url{https://www.youtube.com/watch?v=4FAIiLGCllo&t}, at 1:09:52.} We believe this critique is entirely apt for proposals to simply redact records. But critical to our approach -- and part of AB 1466 -- is the retention of nonredacted instances, making systematic and efficient documentation possible.\footnote{We believe Rose would agree, as she notes, 
``If you erase them from the records, you can’t have anything like these fabulous mapping projects'' (id.).} Technology here, if anything, enabled the understanding and mapping of racial covenants in Santa Clara County. AB 1466 could have mired local governments in compliance initiatives and technology frees up resources to focus on other goals. More generally, California's AB 1466 was a reaction against the perceived inefficacy of the law that required homeowners to take the initiative.  And if such statewide initiatives force greater systemic transparency around the historical use of racial covenants -- when deed records are otherwise only available for purchase on a one-off basis -- we view this as a compelling benefit. 

\section{Conclusion}
\label{sec:conclude}

We have shown that through a unique academic-agency partnership, we were able to prototype, test, and integrate AI to scale the redaction, mapping, and preservation of racial covenants across 5.2 million pages of deed records. Substantively, our approach can empower researchers, governments, and citizens to learn about local histories of discrimination with a level of nuance and specificity that has only been available in a few jurisdictions to date.  For instance, our findings show that while racial covenants in Santa Clara specifically focused on Asian groups, covenants barred Black homeowners at the same rate, even when the Black population was one tenth the Asian population in the county. Consistent with \citep{brooksrose2013saving}, we observe the persistence of racial covenants after 1948. And our finegrained mapping and information extraction enables us to surface specific developers responsible for the bulk of racial covenants, who may have had agency in the construction of Santa Clara County \citep{howell, redford}. 

As over a dozen states have moved to enable the redaction of deed records, a substantial policy choice lies in whether to rely on individual homeowners or government to identify and redact deeds. The main reason for the former has  been the perceived cost. But our collaboration demonstrates that AI systems make more proactive efforts like California's eminently feasible at significantly lower cost than conventionally perceived. The benefits of more proactive efforts are substantial, resulting in a speedier and  much more comprehensive accounting of housing discrimination than piecemeal efforts.    

Our collaboration also paints a promising path for AI in the public sector. 

First, the search challenge for legal reform is pervasive. The U.S.\ Congress struggles with tracking the number of mandated reports strewn about the U.S.\ Code, with many suspected to be obsolete \citep{pray2005congressional}. When California liberalizes approvals for ``accessory dwelling units'' to address the acute housing shortage, harmonization with each county, municipal, and administrative regulation proves taxing. In one municipality, attorneys have referred to the urgent need for ``code cleanup,'' given the many outdated or irrelevant provisions that persist. Herein lies the substantial promise of AI to help in turning what for Los Angeles is an \$8M, 7-year project into a rapid process.

Second, our collaboration also speaks to an increasingly important policy debate around the regulation of open vs.\ closed AI models. Calls for regulation of open models have focused on the potential for risk if developers cannot control their usage via API \citep{pmlr-v235-kapoor24a}. Few projects have provided concrete numbers on the marginal benefits in production use, and our application shows the immense cost savings that can be associated with open models. The cost of an open model is 2\% %
of the comparable cost of a proprietary model.\footnote{This is based on the comparison of costs of the finetuned Mistral model and GPT 3.5 in Table~\ref{tab:cost_comparison}. The cost of the open model is roughly 0.5\% of GPT-4 Turbo.} Given the long history of procurement challenges for government technology \citep{pahlka2023recoding} and the federal government's open source policy\footnote{Office of Management and Budget, Federal Source Code Policy: Achieving Efficiency, Transparency, and Innovation through Reusable and Open Source Software, Aug.\ 8, 2016, \url{https://obamawhitehouse.archives.gov/sites/default/files/omb/memoranda/2016/m_16_21.pdf}.}, this cost finding is of particular significance.  
We release both the fine-tuned language model and the web application to assist in reviewing model outputs to enable jurisdictions to efficiently and effectively explore utilizing these approaches. 

Third, many have called for AI regulation based on a categorical determination of whether AI systems are rights-impacting. The U.S. federal government's response, for instance, triggers a bundle of process-based controls for AI systems that ``serve[] as
a principal basis for a decision or action.''\footnote{Office of Management and Budget, Advancing Governance, Innovation, and Risk Management for Agency Use of Artificial Intelligence, March 28, 2024, \url{https://www.whitehouse.gov/wp-content/uploads/2024/03/M-24-10-Advancing-Governance-Innovation-and-Risk-Management-for-Agency-Use-of-Artificial-Intelligence.pdf}.}  Our project illustrates how the integration of AI models can exist along a spectrum within a government program. AI to improve OCR should be subject to very different safeguards than AI to fully automate racial covenant redaction. Responsible adoption requires identifying the appropriate point of integration, and safeguards must be tailored to risk. If, for instance, any use of AI in OCR triggered the right to opt out of AI systems, government programs will suffer \citep{martin_spectrum_2024}. 

Last, while one of the prevalent anxieties around AI is the potential for irresponsible usage to exacerbate biases -- for which there is an abundance of evidence \citep[e.g.,][]{buolamwini2018gender, liang2021towards} -- our collaboration shows the immense benefits for affirmatively using AI to uncover historical discrimination, promote an improved understanding of pathways for disparities, and reduce the stigmatic and signaling harms from racial covenants.

\section*{Acknowledgements}
We thank Nikita Bhardwaj and Helen Gu for research assistance; Gina Alcomendras, Louis Chiaramonte, Robert Fannion, Greta Hansen, Margaret Pula, Anthony Serafica, and Genevieve Singh-Hanzlick for the collaboration; Ananya Karthik and Allison Casasola for labeling assistance; and Greg Ablavsky, Michael Corey, Kirsten Delegard, Sarah DeMott, Dan Jurafsky, Gideon Lichfield, Varun Magesh, Erin Maneri, Derek Ouyang, Claire Lazar Reich, Carol Rose, Kit Rodolfa, Kyle Swanson, and Andrea Vallebueno for helpful feedback and comments. 

\clearpage
\bibliographystyle{acl_natbib}
\bibliography{references}

\begin{thebibliography}{43}
\providecommand{\natexlab}[1]{#1}

\bibitem[{Bakelmun and Shoenfeld(2019)}]{bakelmun2019open}
Ashley Bakelmun and Sarah~Jane Shoenfeld. 2019.
\newblock Open data and racial segregation: Mapping the historic imprint of racial covenants and redlining on american cities.
\newblock In \emph{Open Cities| Open Data: Collaborative Cities in the Information Era}, pages 57--83. Springer.

\bibitem[{Brooks(2011)}]{brooks2011covenants}
Richard R~W Brooks. 2011.
\newblock Covenants without courts: enforcing residential segregation with legally unenforceable agreements.
\newblock \emph{American Economic Review}, 101(3):360--365.

\bibitem[{Brooks and Rose(2013)}]{brooksrose2013saving}
Richard R.~W. Brooks and Carol~M. Rose. 2013.
\newblock \emph{{Saving the Neighborhood: Racially Restrictive Covenants, Law, and Social Norms}}.
\newblock Harvard University Press.

\bibitem[{Buolamwini and Gebru(2018)}]{buolamwini2018gender}
Joy Buolamwini and Timnit Gebru. 2018.
\newblock Gender shades: Intersectional accuracy disparities in commercial gender classification.
\newblock In \emph{Conference on fairness, accountability and transparency}, pages 77--91. PMLR.

\bibitem[{{California State Legislature}(2021)}]{california_ab1466}
{California State Legislature}. 2021.
\newblock \href {https://legiscan.com/CA/text/AB1466/id/2434855} {{California Assembly Bill No. 1466 (AB 1466)}, \emph{Real property: discriminatory restrictions}}.

\bibitem[{{City Roots Community Land Trust} and {Yale Environmental Protection Clinic}(2020)}]{yalereport}
{City Roots Community Land Trust} and {Yale Environmental Protection Clinic}. 2020.
\newblock \href {https://law.yale.edu/sites/default/files/area/clinic/document/2020.7.31_-_confronting_racial_covenants_-_yale.city_roots_guide.pdf} {{Confronting Racial Covenants: How They Segregated Monroe County and What to Do About Them}}.

\bibitem[{Engstrom et~al.(2020)Engstrom, Ho, Sharkey, and Cuéllar}]{engstrom_government_2020}
David~Freeman Engstrom, Daniel~E. Ho, Catherine Sharkey, and Mariano-Florentino Cuéllar. 2020.
\newblock \href {https://www.acus.gov/sites/default/files/documents/Government%20by%20Algorithm.pdf} {\emph{Government by {Algorithm}: {Artificial} {Intelligence} in {Federal} {Administrative} {Agencies}}}.
\newblock Administrative Conference of the United States.

\bibitem[{Gonda(2015)}]{gonda2015unjust}
Jeffrey~D Gonda. 2015.
\newblock \emph{Unjust deeds: The restrictive covenant cases and the making of the civil rights movement}.
\newblock UNC Press Books.

\bibitem[{Gotham(2000)}]{gotham2000urban}
Kevin~Fox Gotham. 2000.
\newblock Urban space, restrictive covenants and the origins of racial residential segregation in a {US} city, 1900--50.
\newblock \emph{International Journal of Urban and Regional Research}, 24(3):616--633.

\bibitem[{Gray and Suri(2019)}]{gray2019ghost}
Mary~L Gray and Siddharth Suri. 2019.
\newblock \emph{Ghost work: How to stop Silicon Valley from building a new global underclass}.
\newblock Eamon Dolan Books.

\bibitem[{Grier and Grier(1960)}]{taeuber1961privately}
Eunice Grier and George Grier. 1960.
\newblock \emph{Privately Developed Interracial Housing: An Analysis of Experience}.
\newblock University of California Press, Berkeley, CA.

\bibitem[{Halberstadt(1997)}]{halberstadt}
April~Hope Halberstadt. 1997.
\newblock \emph{{The Willow Glen Neighborhood: Then and Now}}.
\newblock Renacsi.

\bibitem[{Howard(2021)}]{howard2021california}
Lexi~Purich Howard. 2021.
\newblock California swings for the fences to strike racially restrictive covenants from the public record.
\newblock \emph{California Real Property Law Journal}, 39(4).

\bibitem[{Howell(2016)}]{howell}
Ocean Howell. 2016.
\newblock \href {https://doi.org/10.1525/phr.2016.85.3.379} {{The Merchant Crusaders: Eichler Homes and Fair Housing, 1949–1974}}.
\newblock \emph{Pacific Historical Review}, 85(3):379--407.

\bibitem[{Hu et~al.(2021)Hu, Shen, Wallis, Allen-Zhu, Li, Wang, Wang, and Chen}]{hu_lora_2021}
Edward~J. Hu, Yelong Shen, Phillip Wallis, Zeyuan Allen-Zhu, Yuanzhi Li, Shean Wang, Lu~Wang, and Weizhu Chen. 2021.
\newblock \href {https://doi.org/10.48550/arXiv.2106.09685} {{LoRA}: {Low}-{Rank} {Adaptation} of {Large} {Language} {Models}}.
\newblock \emph{arXiv preprint}.
\newblock ArXiv:2106.09685 [cs].

\bibitem[{Jiang et~al.(2023)Jiang, Sablayrolles, Mensch, Bamford, Chaplot, Casas, Bressand, Lengyel, Lample, Saulnier, Lavaud, Lachaux, Stock, Scao, Lavril, Wang, Lacroix, and Sayed}]{jiang_mistral_2023}
Albert~Q. Jiang, Alexandre Sablayrolles, Arthur Mensch, Chris Bamford, Devendra~Singh Chaplot, Diego de~las Casas, Florian Bressand, Gianna Lengyel, Guillaume Lample, Lucile Saulnier, Lélio~Renard Lavaud, Marie-Anne Lachaux, Pierre Stock, Teven~Le Scao, Thibaut Lavril, Thomas Wang, Timothée Lacroix, and William~El Sayed. 2023.
\newblock \href {https://doi.org/10.48550/ARXIV.2310.06825} {Mistral {7B}}.
\newblock \emph{arXiv preprint}.
\newblock Version Number: 1.

\bibitem[{Jones-Correa(2000)}]{jones2000origins}
Michael Jones-Correa. 2000.
\newblock The origins and diffusion of racial restrictive covenants.
\newblock \emph{Political Science Quarterly}, 115(4):541--568.

\bibitem[{Kapoor et~al.(2024)Kapoor, Bommasani, Klyman, Longpre, Ramaswami, Cihon, Hopkins, Bankston, Biderman, Bogen, Chowdhury, Engler, Henderson, Jernite, Lazar, Maffulli, Nelson, Pineau, Skowron, Song, Storchan, Zhang, Ho, Liang, and Narayanan}]{pmlr-v235-kapoor24a}
Sayash Kapoor, Rishi Bommasani, Kevin Klyman, Shayne Longpre, Ashwin Ramaswami, Peter Cihon, Aspen~K Hopkins, Kevin Bankston, Stella Biderman, Miranda Bogen, Rumman Chowdhury, Alex Engler, Peter Henderson, Yacine Jernite, Seth Lazar, Stefano Maffulli, Alondra Nelson, Joelle Pineau, Aviya Skowron, Dawn Song, Victor Storchan, Daniel Zhang, Daniel~E. Ho, Percy Liang, and Arvind Narayanan. 2024.
\newblock \href {https://proceedings.mlr.press/v235/kapoor24a.html} {Position: On the societal impact of open foundation models}.
\newblock In \emph{Proceedings of the 41st International Conference on Machine Learning}, volume 235 of \emph{Proceedings of Machine Learning Research}, pages 23082--23104. PMLR.

\bibitem[{Kelman(1990)}]{kelman1990procurement}
Steve Kelman. 1990.
\newblock Procurement and public management.
\newblock Technical report, American Enterprise Institute.

\bibitem[{Liang et~al.(2021)Liang, Wu, Morency, and Salakhutdinov}]{liang2021towards}
Paul~Pu Liang, Chiyu Wu, Louis-Philippe Morency, and Ruslan Salakhutdinov. 2021.
\newblock Towards understanding and mitigating social biases in language models.
\newblock In \emph{International Conference on Machine Learning}, pages 6565--6576. PMLR.

\bibitem[{Liao et~al.(2020)Liao, Wan, Yao, Chen, and Bai}]{liao2020real}
Minghui Liao, Zhaoyi Wan, Cong Yao, Kai Chen, and Xiang Bai. 2020.
\newblock Real-time scene text detection with differentiable binarization.
\newblock In \emph{Proceedings of the AAAI conference on artificial intelligence}, volume~34, pages 11474--11481.

\bibitem[{Magesh et~al.(2024)Magesh, Surani, Dahl, Suzgun, Manning, and Ho}]{magesh2024hallucination}
Varun Magesh, Faiz Surani, Matthew Dahl, Mirac Suzgun, Christopher~D Manning, and Daniel~E Ho. 2024.
\newblock Hallucination-free? {A}ssessing the reliability of leading {AI} legal research tools.
\newblock \emph{arXiv preprint arXiv:2405.20362}.

\bibitem[{{Mapping Prejudice Project}(2022)}]{MappingPrejudice2022_RRCs}
{Mapping Prejudice Project}. 2022.
\newblock \href {https://mappingprejudice.umn.edu/racial-covenants/what-is-a-covenant} {{What is a Covenant?}}

\bibitem[{Martin et~al.(2024)Martin, Surani, Rodolfa, Perez, and Ho}]{martin_spectrum_2024}
Olivia Martin, Faiz Surani, Kit Rodolfa, Amy Perez, and Daniel~E. Ho. 2024.
\newblock The {Spectrum} of {AI} {Integration}: {The} {Case} of {Benefits} {Adjudication}.
\newblock In Cynthia~H. Cwik, Christopher~A. Suarez, and Lucy~L. Thomson, editors, \emph{{AI}: {Legal} {Issues}, {Policy}, and {Practical} {Strategies}}. American Bar Association.

\bibitem[{Mindee(2021)}]{doctr2021}
Mindee. 2021.
\newblock doctr: Document text recognition.
\newblock \url{https://github.com/mindee/doctr}.

\bibitem[{Ming(1949)}]{ming1949racial}
William~R Ming. 1949.
\newblock Racial restrictions and the fourteenth amendment: The restrictive covenant cases.
\newblock \emph{The University of Chicago Law Review}, 16(2):203--238.

\bibitem[{Pahlka(2023)}]{pahlka2023recoding}
Jennifer Pahlka. 2023.
\newblock \emph{Recoding America: why government is failing in the digital age and how we can do better}.
\newblock Metropolitan Books.

\bibitem[{Papineni et~al.(2002)Papineni, Roukos, Ward, and Zhu}]{papineni2002bleu}
Kishore Papineni, Salim Roukos, Todd Ward, and Wei-Jing Zhu. 2002.
\newblock Bleu: a method for automatic evaluation of machine translation.
\newblock In \emph{Proceedings of the 40th annual meeting of the Association for Computational Linguistics}, pages 311--318.

\bibitem[{Pray(2005)}]{pray2005congressional}
Jonathan~G Pray. 2005.
\newblock Congressional reporting requirements: Testing the limits of the oversight power.
\newblock \emph{U. Colo. L. Rev.}, 76:297.

\bibitem[{Redford(2017)}]{redford}
Laura Redford. 2017.
\newblock \href {https://doi.org/10.1177/1538513216676191} {The intertwined history of class and race segregation in los angeles}.
\newblock \emph{Journal of Planning History}, 16(4):305--322.

\bibitem[{Rhoads(2018)}]{rhoads_cemetery}
Loren Rhoads. 2018.
\newblock \href {https://cemeterytravel.com/2018/05/19/cemetery-of-the-week-167-oak-hill-memorial-park/} {Cemetery of the {Week} \#167: {Oak} {Hill} {Memorial} {Park}}.

\bibitem[{Rohrbach et~al.(2018)Rohrbach, Hendricks, Burns, Darrell, and Saenko}]{rohrbach2018object}
Anna Rohrbach, Lisa~Anne Hendricks, Kaylee Burns, Trevor Darrell, and Kate Saenko. 2018.
\newblock Object hallucination in image captioning.
\newblock \emph{arXiv preprint arXiv:1809.02156}.

\bibitem[{Roisman(2022)}]{roisman2022stumbling}
Florence~Wagman Roisman. 2022.
\newblock Stumbling stones at levittown: What to do about racial covenants in the united states.
\newblock \emph{Journal of Affordable Housing Volume}, 30(3).

\bibitem[{Rose(2022)}]{rose2022property}
Carol~M Rose. 2022.
\newblock Property law and inequality: Lessons from racially restrictive convenants.
\newblock \emph{Nw. UL Rev.}, 117:225.

\bibitem[{Rose(2024)}]{rose2023general}
Carol~M Rose. 2024.
\newblock General customs and legal institutions: The short, sad example of racially restrictive covenants in the {United States}.
\newblock In Samuel~L. Bray, John~CP Goldberg, Paul~B. Miller, and Henry~E. Smith, editors, \emph{Interstitial Private Law}. Oxford University Press.

\bibitem[{Rose et~al.(2016)Rose, Brooks, Brown, and Smith}]{rose2016racial}
Carol~M Rose, Richard~RW Brooks, A~Brown, and V~Smith. 2016.
\newblock {Racial Covenants and Housing Segregation, Yesterday and Today}.
\newblock \emph{Race and Real Estate}, pages 161--76.

\bibitem[{Rothstein(2017)}]{rothstein2017color}
Richard Rothstein. 2017.
\newblock \emph{{The Color of Law: A Forgotten History of How Our Government egregated America}}.
\newblock Liveright Publishing.

\bibitem[{Santucci(2020)}]{santucci2020documenting}
Larry Santucci. 2020.
\newblock {Documenting Racially Restrictive Covenants in 20th Century Philadelphia}.
\newblock \emph{Cityscape}, 22(3):241--268.

\bibitem[{Simonyan and Zisserman(2014)}]{simonyan2014very}
Karen Simonyan and Andrew Zisserman. 2014.
\newblock Very deep convolutional networks for large-scale image recognition.
\newblock \emph{arXiv preprint arXiv:1409.1556}.

\bibitem[{{US Census Bureau}(1942)}]{us_census_bureau_1940_nodate}
{US Census Bureau}. 1942.
\newblock \href {https://www.census.gov/library/publications/1942/dec/population-vol-1.html} {1940 {Census} of {Population}: {Volume} 1. {Number} of {Inhabitants}. {Total} {Population} for {States}, {Counties}, and {Minor} {Civil} {Divisions}; for {Urban} and {Rural} {Areas}; for {Incorporated} {Places}; for {Metropolitan} {Districts}; and for {Census} {Tracts}}.
\newblock Section: Government.

\bibitem[{Vose(1967)}]{vose1967caucasians}
Clement~E Vose. 1967.
\newblock \emph{Caucasians only: The Supreme Court, the NAACP, and the restrictive covenant cases}.
\newblock Univ of California Press.

\bibitem[{Wagstaff(2012)}]{wagstaff2012machine}
KL~Wagstaff. 2012.
\newblock Machine learning that matters.
\newblock In \emph{Proceedings of the 29th International Conference on Machine Learning, 2012}, pages 529--536.

\bibitem[{Wang et~al.(2020)Wang, Yao, Kwok, and Ni}]{wang2020generalizing}
Yaqing Wang, Quanming Yao, James~T Kwok, and Lionel~M Ni. 2020.
\newblock Generalizing from a few examples: A survey on few-shot learning.
\newblock \emph{ACM computing surveys (csur)}, 53(3):1--34.

\end{thebibliography}

\clearpage
\appendix
\section{Ethical Statement}

We here discuss our considerations for the use and integration of machine learning in this context. 

First, the AB 1466 setting is one where resource constraints loom large. Purely human review would make implementation of AB 1466 challenging and funding cannot fully support such purely manual efforts (see Section~\ref{sec:background}). Our work helps prioritize available human and attorney resources to documents with high likelihood of racial covenants. This makes the process of redaction faster---meaning that fewer homeowners will have to sign such offensive covenants in transactions or expend their own resources in removing them.

Second, AB 1466 requires human oversight in the form of counsel review of any proposed redaction (see Figure~\ref{fig:integration}). The risk of false positives is hence primarily about the review time required. There is a risk that our system misses racial covenants, but (a) it is not obvious whether human reviewers would fare better given the high recall performance reported in Section~\ref{sec:evaluation}, and (b) AB 1466 provides alternative mechanisms to flag remaining racial covenants.\footnote{For instance, AB 1466 places obligations on a title company, escrow company, real estate broker, real estate agent, or association to notify and assist homeowners with redaction. Cal.\ Gov.\ Code \S~12956.2.}

Third, there are substantial societal benefits from reducing the stigmatic and signaling effects of racial covenants and enabling a systematic historical accounting of racial covenants and understanding of mechanisms of housing discrimination. 

Fourth, the release of the fine-tuned Mistral model does not pose any marginal risk (since Mistral is already an open model), while posing substantial benefits for the many jurisdictions grappling with efforts to identify and redact racial covenants. While we have curated a diverse training dataset from eight counties across the United States, we do caution that application in other jurisdictions should include domain-specific validation efforts. Furthermore, there are no additional privacy risks from releasing this model, we mask the loss on input text (so a model cannot learn to generate anything in its input) and we verified that training data for the covenant itself does not contain private information.

Fifth, we were guided in all elements of this collaboration by the articulated needs from County partners, when government entities have historically faced challenges integrating new technology \citep{kelman1990procurement}. Our approach centered the concrete problem faced by administrators to ensure that machine learning was appropriately prototyped, developed, and integrated (see Section~\ref{ref:integration}). Such an approach requires support and cooperation with the county, but may not be feasible in all counties. As such, community-based volunteer efforts across the country will likely still play an important role, and such efforts may also be enhanced by our model. 

In short, due to minimal risks and substantial benefits, we believe this to be a compelling illustration of machine learning for public good \citep{wagstaff2012machine}. 

\section{Optical Character Recognition (OCR) Experiments}
\label{appendix_ocr}

\begin{figure}[h]
    \centering
    \includegraphics[width=0.8\linewidth]{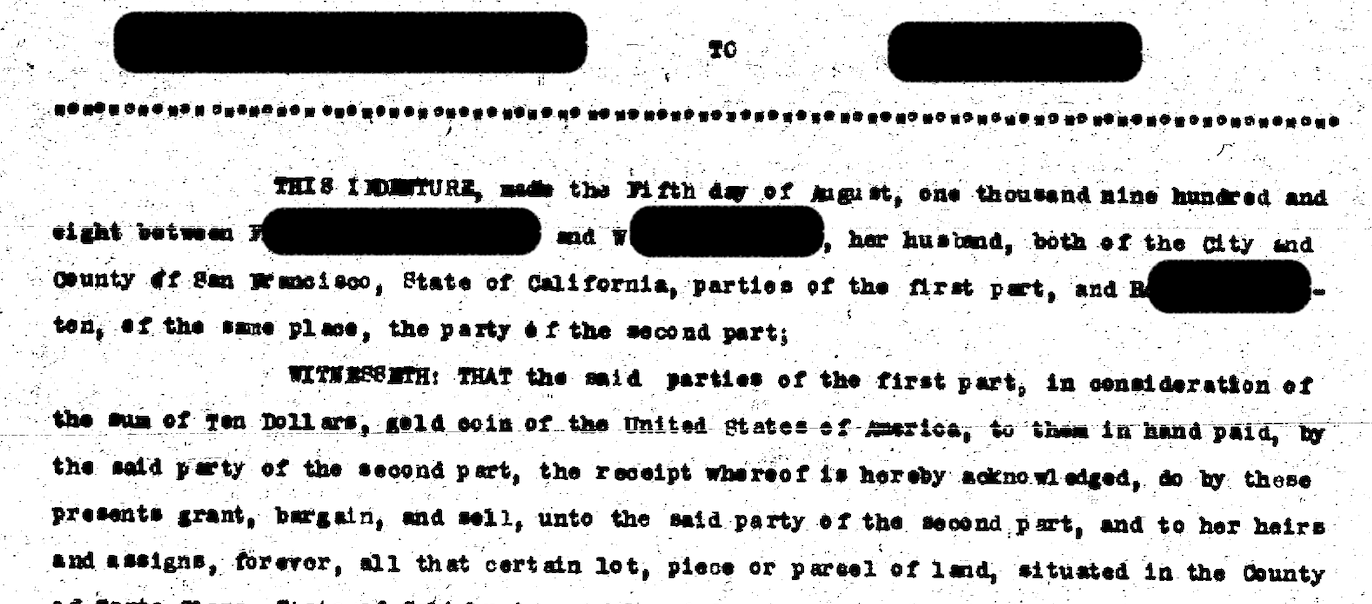}
    \caption{Excerpt from a 1908 property deed scan, illustrating challenges for OCR: low resolution, smudged characters, and visual artifacts.}
    \label{fig:ocr-challenges}
    \vspace{0.5em}
\end{figure}

Historical property deeds prior to the late 20th century are typically typewritten and vary widely in text and scan quality. As such, OCR is a non-trivial problem for these documents.

In our experiments, we found that the most popular open-source OCR library, Tesseract, performed very poorly on older deeds, particularly those with low-quality scans. Tesseract transcriptions often omitted entire words, substituted characters, and misordered sections of text. We found that common spell-correction and other off-the-shelf methods were insufficient to mitigate these issues.

However, the DocTR library with the ResNet-50 and VGG-16 models is more effective. The improvements in OCR after switching to DocTR further lead to downstream increases in the performance of our RRC detection models. We qualitatively and quantitatively compare the performance of the outputs of both OCR libraries on our dataset.

\begin{table}[h]
\centering
\begin{tabular}{lll}
\hline
 & \multicolumn{2}{l}{\textbf{F1 Score on Evaluation Set}} \\
\textbf{Detector} & \textbf{Tesseract} & \textbf{DocTR: ResNet-50 / VGG-16} \\
\hline
\textit{Fuzzy Matching}& 0.921 & \textbf{0.943} \\
\textit{Mistral Fine-Tuned}& 0.986 & \textbf{0.997} \\
\hline
\end{tabular}
\caption{Switching from Tesseract to DocTR OCR improved model performance for both the baseline and the language models.}
\end{table}

\textbf{Tesseract Example}
    
; an THIS. 1DENTUE, ‘made ‘ott day of February in the ; year of) ar Lord ‘ane. “| thoamnd nine ‘hundred. and ten, “by and. ve tween [REDACTED] of santa clara. county, state. "of california; the party ‘of the first part, and wre. [REDACTED], ‘of san mec county, | State of ouitornia, the party of the second part; — aa re se ese Mi TsESSETH: THAT the emiaiparty of the fitet ‘part; for and in consideration: : Ms of. tte, ain of: ‘ten (gio) Dollars, eae éoin of the ‘United: ‘Btates ‘to. hin in hand paid. by the os ‘gata party ‘er the second pent, the receipt whereof ie herety scimamtedged, has ‘granted, are - : gained and so1,, conveyed and confirma, and “by these presents dos. (grant,: vargain and sell, “SE

\textbf{DocTR Example}

THIS INDETURE, made the7th day of rebruary in the year of aur Lord thousand nine hundred and ten, by and. be tween [REDACTED] a Santa Clara. County, Stateof califarnia, the. party of the first part, and [REDACTED], of San iteo county,of tlie, sum of Tep (\$10) Dallars, gold coin, of the United Statez to.hin in hand paid. by thebaid party'ar the second pant, the receipt Whereof 10 herehy aeknaledged, has granted, bar-gained and sol4,. conveyed and confirmed, and by these presents dos grant, bargain and sell,convey ad confirn anto the said party of the second. part, and to her beireiand assigna forever

\begin{figure}[!t]
    \centering
    \setlength{\belowcaptionskip}{10pt}
    \includegraphics[width=\linewidth]{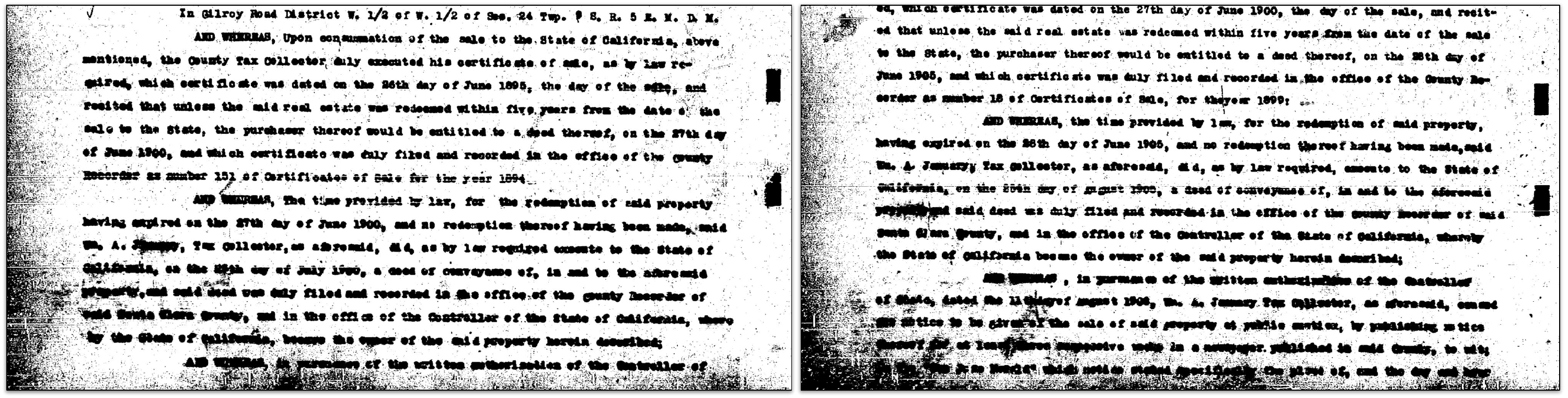}
    \caption[]{Real property deeds from Santa Clara County (from 1908), with several visible scanning artifacts.}
    \label{fig:oct-example}
\end{figure}

\clearpage
\section{Instruction-Finetuning Prompt}
\label{sec:instruction-finetuning-prompt}

\begin{figure}[!h]
    \centering
    \setlength{\belowcaptionskip}{10pt}
    \includegraphics[width=\linewidth]{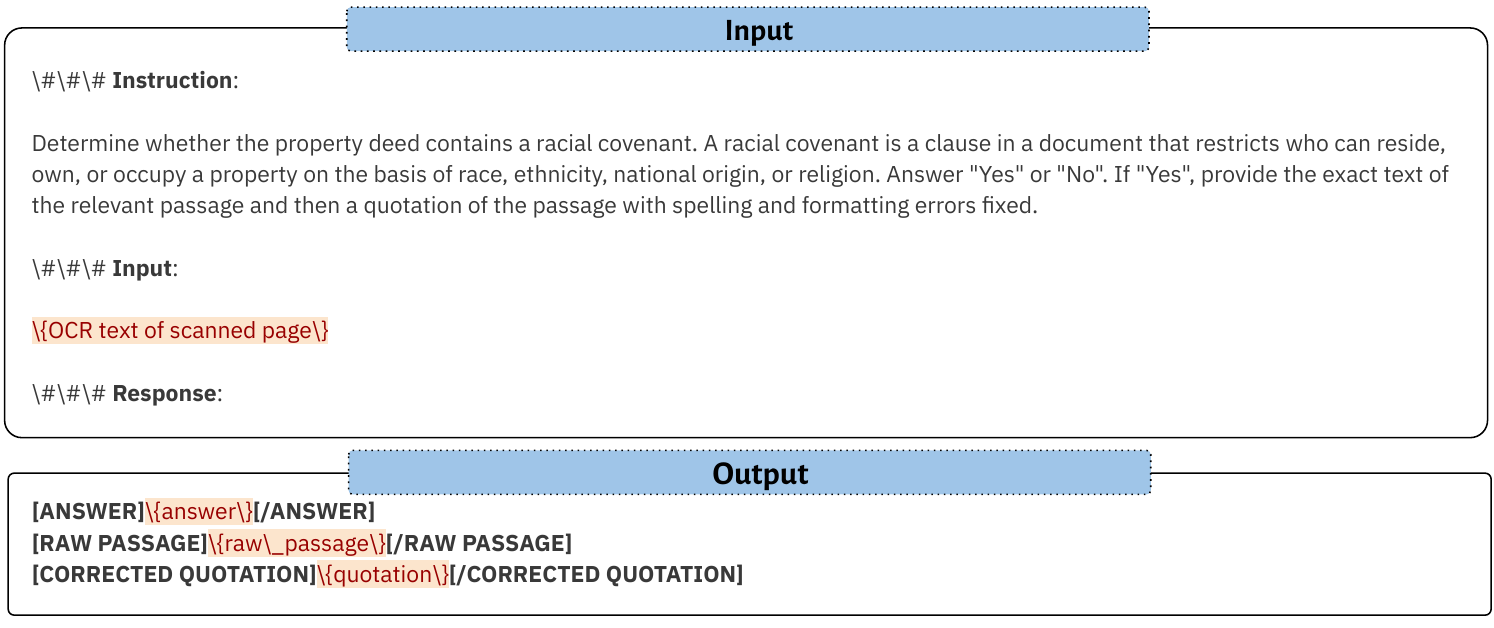}
    \caption[]{Input and output template used in the finetuning of our Mistral 7B model.}
    \label{fig:instruction-finetuning-template}
\end{figure}

\section{Annotation App}

\begin{figure}[h]
    \centering
    \includegraphics[width=0.8\linewidth]{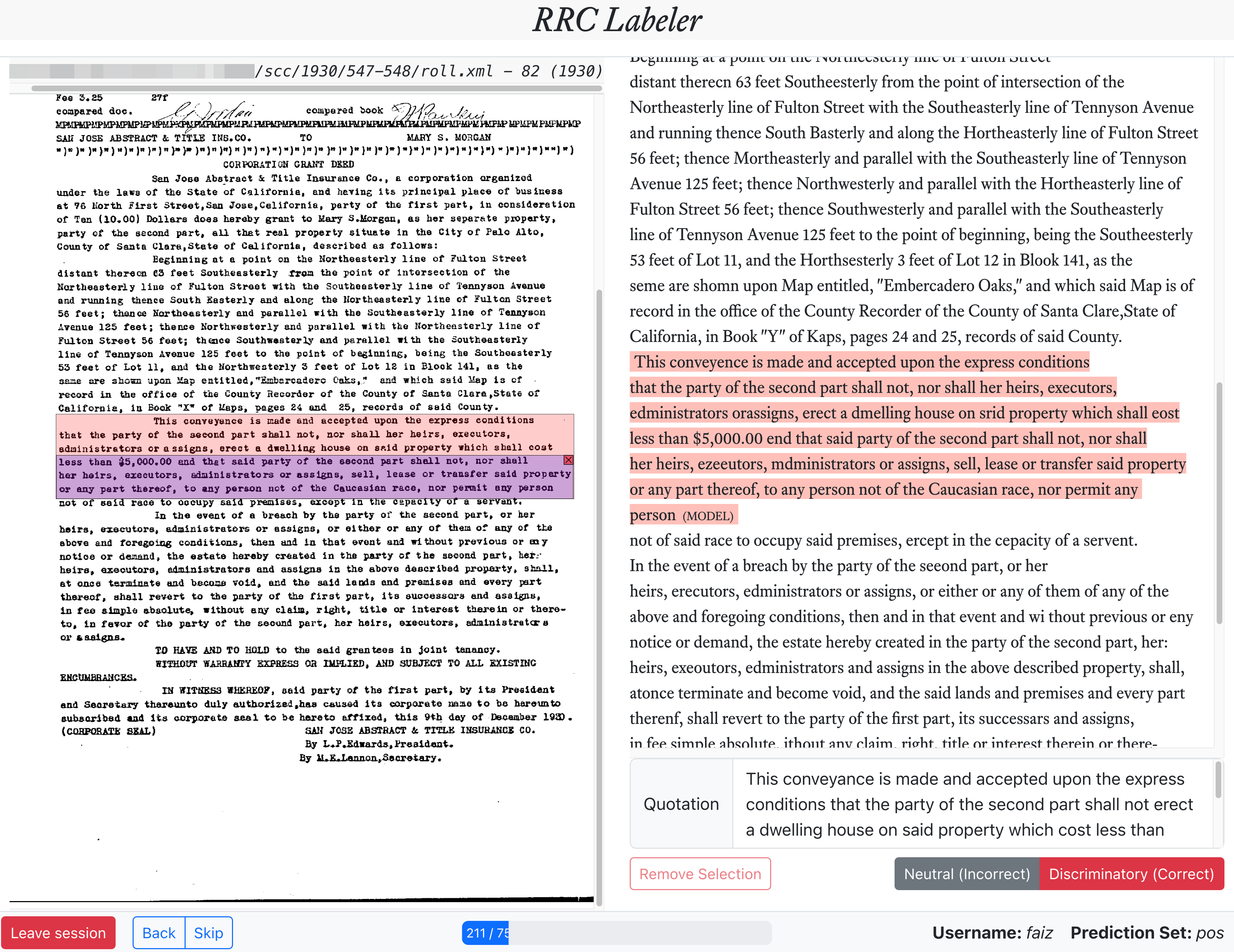}
    \caption{Screenshot of the app used to label data for training our racial covenant detectors. The interface allows labelers to modify or reverse the initial annotation of the model with respect to the on-page bounding box, text span, and cleaned up quotation.}
    \label{fig:labeler_app}
    \vspace{0.5em}
\end{figure}

\clearpage
\section{Resource Cost Comparison}
\label{sec:resource-cost-comparison}

\textbf{Manual Review.} Manual review was the first -- and most straightforward yet least feasible -- option considered by Santa Clara County for implementing the review of racial covenants in property records. With an estimated review speed of 60 pages per hour, manually reviewing the entire archive of 84 million pages would require approximately 1.4 million staff hours, or about 160 years of continuous work for a single individual. At the current minimum wage in California (\$16 per hour), the total labor cost for reviewing all 84 million pages would amount to \$22.4 million.

For the subset of 5.2 million pages, manual review would require 86,667 staff hours, resulting in a labor cost of approximately \$1.4 million. These estimates, however, do not account for additional costs, such as staff training, workflow management, and handling the physical or digital logistics associated with the records. Moreover, the extended duration of manual review increases the risk of non-compliance with legislative deadlines, such as those set by California's AB 1466, further emphasizing the impracticality of this approach.

\textbf{Off-the-Shelf Language Model (ChatGPT).} A more technologically advanced and reliable alternative to manual review is the use of a commercial off-the-shelf language model, such as OpenAI’s ChatGPT-3.5, to process property deeds. With an estimated processing capacity of 1,000 requests per minute, analyzing the 5.2 million pages would take about 87 hours. However, while obviously way faster than manual review, the financial cost of using a commercial language model is significant. Based on a token usage of 922 tokens per deed page -- accounting for both the task instructions and document content\footnote{This estimate is derived from a random sample of 10,000 pages from Santa Clara County’s real property records.} -- and OpenAI’s current pricing of \$1 per 1 million tokens for ChatGPT-3.5 (viz., \texttt{gpt-3.5-turbo-1106}), the total cost of processing all 5.2 million pages using a zero-shot prompting setup would be approximately \$4,794.

If a few-shot prompting technique is applied to enhance the model’s accuracy—by providing two examples with each request—the cost rises to \$13,634. For more complex and nuanced tasks like this, a more powerful model, such as GPT-4 Turbo, might be required. At \$10 per million tokens, the cost of using GPT-4 Turbo to process the entire dataset would escalate to \$47,944, making it significantly more expensive than our finetuned Mistral model, which achieves the same outcome for just \$258. 

\textbf{Our Finetuned Language Model (Mistral).} The most cost-effective option is our custom finetuned language model approach. Our Mistral model, finetuned specifically on a set of restrictive covenants, can process approximately 1 million pages per day, allowing it to complete the review of 5.2 million pages in just six days. The primary advantage of this approach lies in its cost-efficiency. Rather than relying on a commercial API, the model can be run on rented GPUs, with a rental cost of roughly \$258 for the entire six-day processing period.\footnote{In our case, we made use of Stanford’s own Sherlock compute cluster to conduct our experiments.} Moreover, performing all our experiments and analyses internally eliminates any privacy-related risks, as we retain complete oversight of the data. This means we do not have to rely on external providers, ensuring that sensitive information is handled securely and in compliance with privacy regulations throughout the entire process.

Thus, the total cost for reviewing 5.2 million pages using this custom model would amount to only \$258 -- dramatically lower than the \$1.4 million required for manual review and significantly less than the \$30,000 projected for using an off-the-shelf LLM. While setting up and fine-tuning the model requires some initial effort, this method offers the most scalable, efficient, and cost-effective solution for identifying RRCs in Santa Clara County’s property records.

\clearpage

\section{List of Terms Used by Santa Clara County in Manual Review of Deeds}
\label{sec:scc-keyword-list}

{\color{red} Warning: Due to the nature of the racial covenants, the search terms contain offensive terms.}

The Santa Clara County Clerk-Recorder’s Office initially used the following list of keywords to identify instances of racial covenants in their digitized real property deeds: African, American Asiatic, Aryans, Asian, Asiatic, Black, Blood, Brown, Caucasian, Chauffeurs \emph{(exception: dependent on the context)}, Chinese, Clover, Color, Dago, Domestic Servants, Domiciled, Dyke, Ethiopians, Foreigners, Gandhi, Gardeners (exception: dependent on the context), Gay, Ginzo, Greaser, Hebrews, Hindu, Immigrant, Indian, Interracial, Italian or Italians, Japanese, Jew or Jews, Korean, Lineage, Malays, Master, Mixed race, Mongolian, Native of the Turkish Empire, Negro, Nigga, Nigger, People, Portuguese, Race, Religion, Restricted District, Servants, Turkish, White, Mulatto

\clearpage
\section{Geographic Distribution of Racial Covenants}\label{appendix:full_faceted_map}

\begin{figure}[ht]
    \centering
    \includegraphics[width=\linewidth]{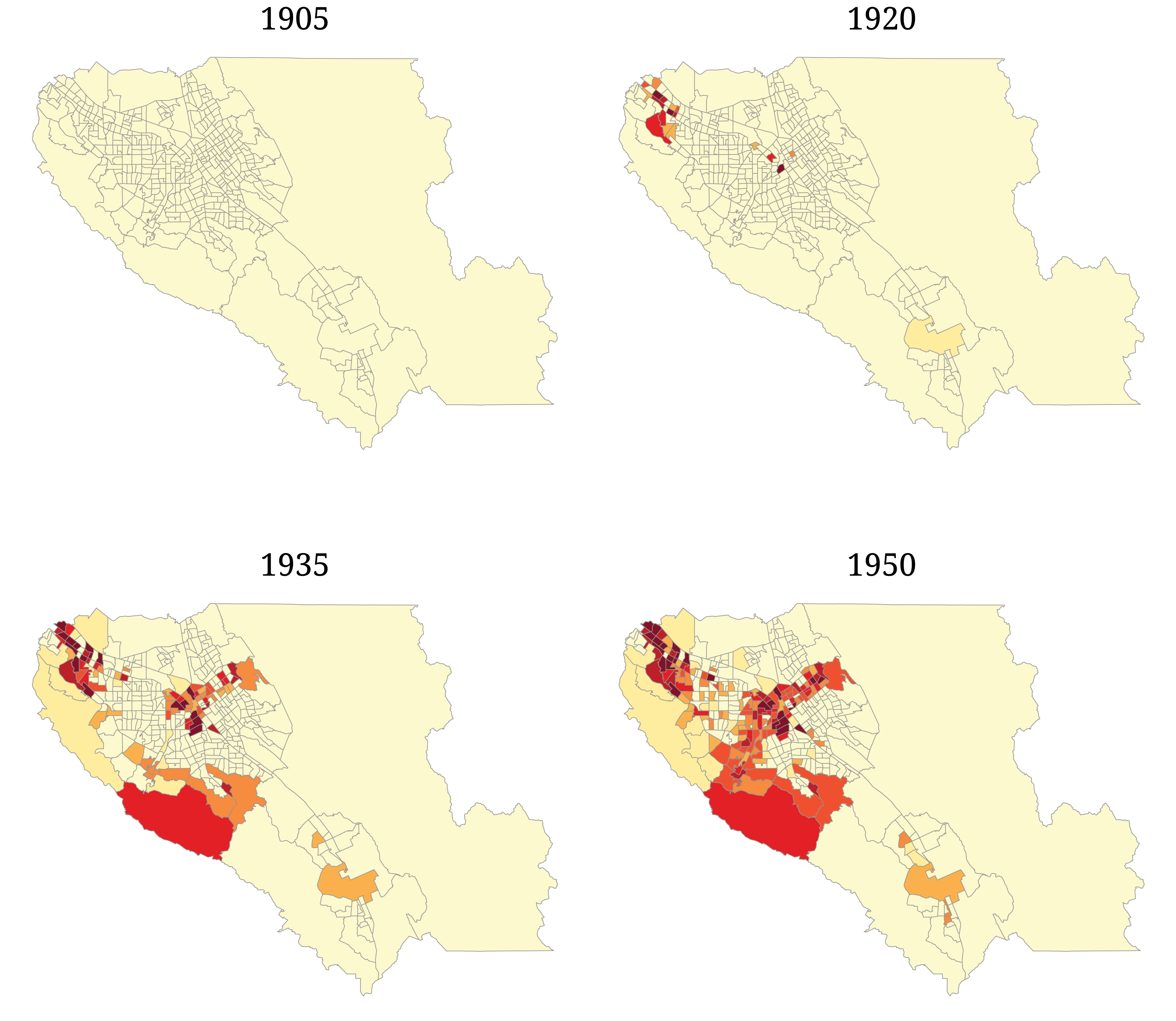}
    \caption[]{Density of properties with racial covenants in modern-day Census tracts in 1905, 1920, 1935, and 1950. This figure includes the more rural tracts in the south and east of the County.}
    \label{fig:scc_map_faceted_full}
\end{figure}

\clearpage
\section{Lot Coverage of Racial Covenants}
\label{appendix:lot-coverage}

This section provides additional visualizations of the distribution of racial covenants in Santa Clara County, focusing on the number of \emph{lots} covered by racial covenants rather than the number of deed records. These figures complement the analysis presented in \S~\ref{sec:prevalence}.

\begin{figure}[h]
    \centering
    \includegraphics[width=\linewidth]{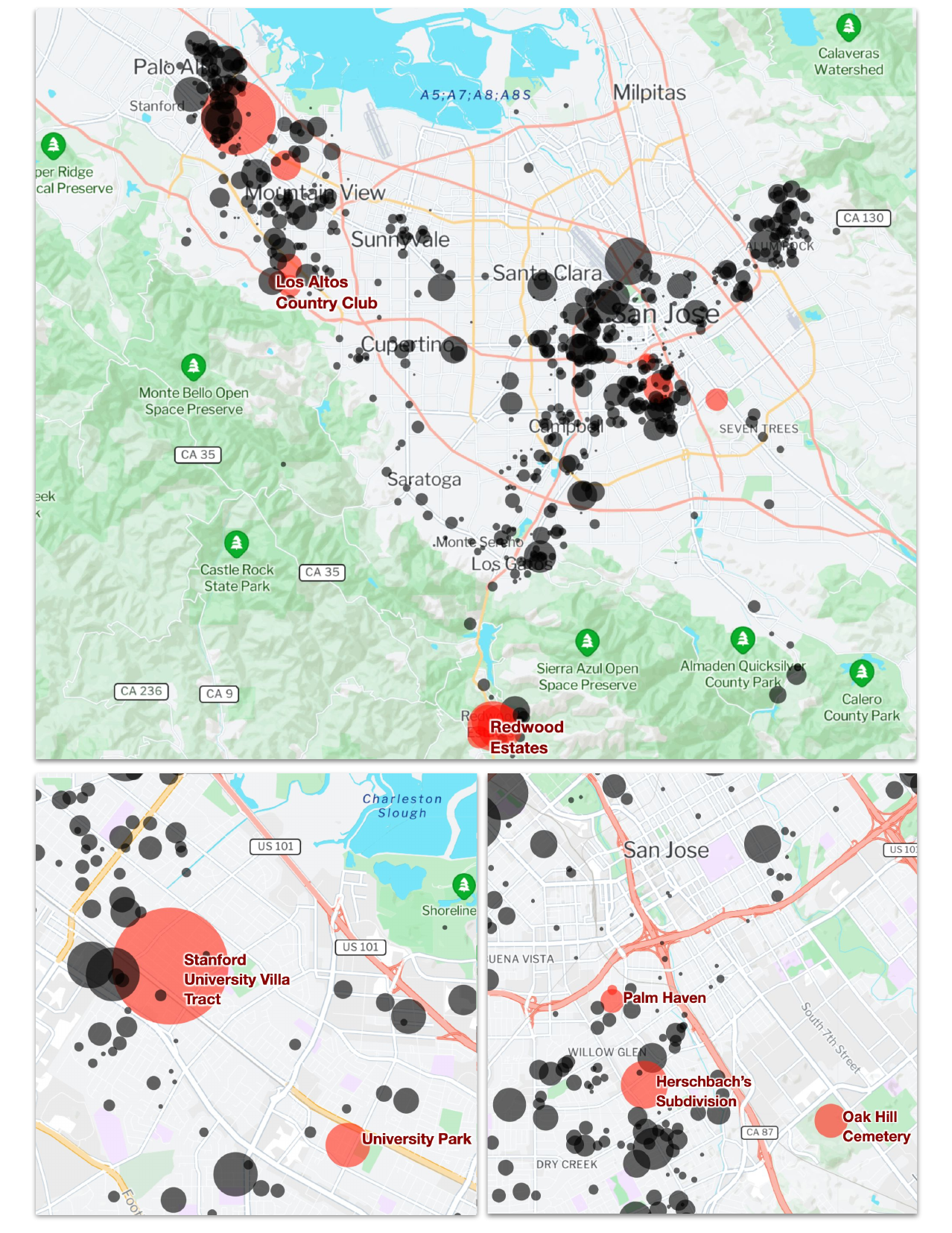}
    \caption{Distribution of lots covered by racial covenants in Santa Clara County. \textbf{Top:} Overview of the entire county. \textbf{Bottom left:} Racial covenants in south Palo Alto and Mountain View. \textbf{Bottom right:} Racial covenants in downtown San Jose. Dots represent individual subdivisions and are scaled in proportion to the number of lots covered by racial covenants, instead of the number of racial covenants.}
    \label{fig:scc_lot_map}
\end{figure}

\begin{figure}[h]
    \centering
    \includegraphics[width=\linewidth]{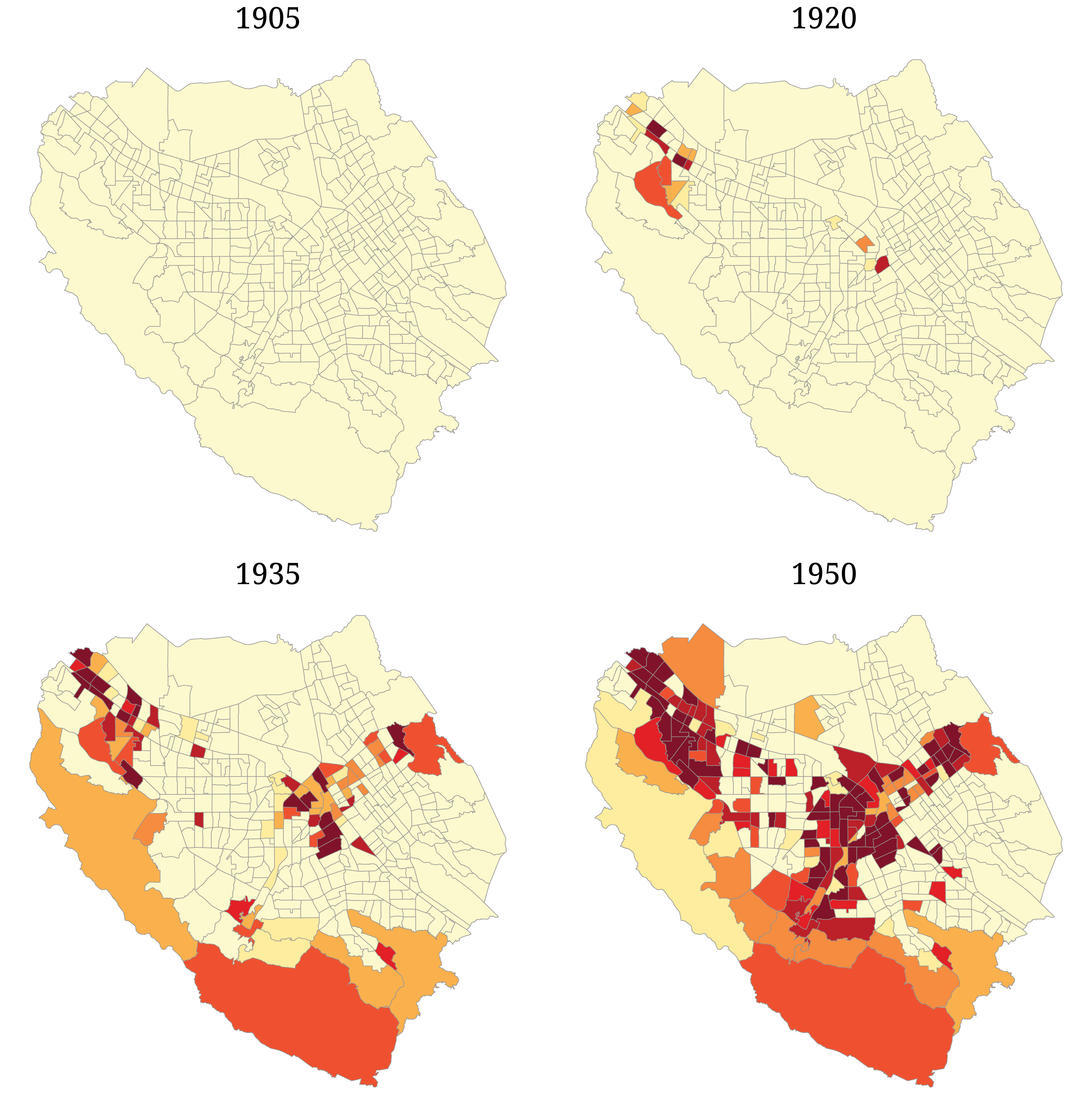}
    \caption{Density of lots covered by racial covenants in modern-day Census tracts in 1905, 1920, 1935, and 1950. Lots are plotted cumulatively. This figure shows the spread of racial covenants throughout the county between 1905 and 1950, with the number of affected lots as the unit of analysis.}
    \label{fig:scc_map_faceted_lot_level}
\end{figure}

\end{document}